\documentclass[lettersize,journal]{IEEEtran}
\usepackage{amsmath,amsfonts}
\usepackage{algorithmic}
\usepackage{algorithm}
\usepackage{array}
\usepackage[caption=false,font=normalsize,labelfont=sf,textfont=sf]{subfig}
\usepackage{textcomp}
\usepackage{stfloats}
\usepackage{url}
\usepackage{verbatim}
\usepackage{graphicx}
\usepackage{cite}
\usepackage{algorithm}  
\usepackage[ruled,vlined,algo2e]{algorithm2e}  
\usepackage{pifont}  
\usepackage{booktabs}
\usepackage{caption}
\usepackage{colortbl} 
\usepackage{makecell}
\usepackage{array}
\usepackage{multirow} 
\usepackage{tabularx}
\usepackage[colorlinks=true, linkcolor=blue, citecolor=blue, urlcolor=blue]{hyperref}
\usepackage{graphicx} 
\usepackage{amssymb}
\usepackage{xcolor}
\usepackage{orcidlink} 
\hypersetup{
    pdfborder={0 0 0}
}
\definecolor{ngreen}{RGB}{17, 173, 30}

\begin{document}

\title{Sculpting Margin Penalty: Intra-Task Adapter Merging and Classifier Calibration for Few-Shot Class-Incremental Learning}

\author{Liang Bai, Hong Song, Jinfu Li, Yucong Lin, Jingfan Fan, Tianyu Fu, Danni Ai, Deqiang Xiao, Jian Yang
\thanks{
Corresponding authors: Hong Song and Jian Yang.
}

\thanks{Liang Bai and Hong Song are with the School of Computer Science and Technology, Beijing Institute of Technology, Beijing 100081, China.}

\thanks{ Jinfu Li, Yucong Lin, Jingfan Fan, Danni Ai, Deqiang Xiao, and Jian Yang are with the School of Optics and Photonics, Beijing Institute of Technology, Beijing 100081, China.}

\thanks{Tianyu Fu is with the School of Medical Technology, Beijing Institute of Technology, Beijing 100081, China.}
}

\markboth{Journal of \LaTeX\ Class Files,~Vol.~14, No.~8, August~2021}%
{Shell \MakeLowercase{\textit{et al.}}: A Sample Article Using IEEEtran.cls for IEEE Journals}


\maketitle

\begin{abstract}
Real-world applications often face data privacy constraints and high acquisition costs, making the assumption of sufficient training data in incremental tasks unrealistic and leading to significant performance degradation in class-incremental learning. Forward-compatible learning, which prospectively prepares for future tasks during base task training, has emerged as a promising solution for Few-Shot Class-Incremental Learning (FSCIL). However, existing methods still struggle to balance base-class discriminability and new-class generalization. Moreover, limited access to original data during incremental tasks often results in ambiguous inter-class decision boundaries. To address these challenges, we propose SMP (Sculpting Margin Penalty), a novel FSCIL method that strategically integrates margin penalties at different stages within the parameter-efficient fine-tuning paradigm. Specifically, we introduce the Margin-aware Intra-task Adapter Merging (MIAM) mechanism for base task learning. MIAM trains two sets of low-rank adapters with distinct classification losses: one with a margin penalty to enhance base-class discriminability, and the other without margin constraints to promote generalization to future new classes. These adapters are then adaptively merged to improve forward compatibility. For incremental tasks, we propose a Margin Penalty-based Classifier Calibration (MPCC) strategy to refine decision boundaries by fine-tuning classifiers on all seen classes' embeddings with a margin penalty. Extensive experiments on CIFAR100, ImageNet-R, and CUB200 demonstrate that SMP achieves state-of-the-art performance in FSCIL while maintaining a better balance between base and new classes.


\end{abstract}


\section{Introduction}


Real-world scenarios are dynamic and continuously introduce new data, posing significant challenges for trained visual recognition models to progressively evolve. Class-Incremental Learning (CIL) \cite{cls_surv,cls_surv2,cil_vt_lw} has emerged as a promising solution by enabling previously trained classification models to incrementally incorporate new classes. CIL research predominantly focuses on designing additional constraints that transfer knowledge from previously learned tasks to guide the learning of new classes, regardless of whether data from previous tasks is available. This backward-compatible approach, which incorporates prior tasks during subsequent learning, facilitates the continuous integration of new concepts while mitigating catastrophic forgetting of previously acquired knowledge. With ongoing advancements in this field, current CIL methods \cite{cil_pass,cil_ssre,cil_praka,cil_vt_rm,cil_vt_mix,cil_tass,acmap} have demonstrated significant performance improvements.

However, these methods typically assume that sufficient training data is available for each incremental task. In practical applications, this assumption often fails due to factors such as data privacy constraints and high acquisition costs. For example, the scarcity of image data for rare wildlife species presents a significant obstacle to building effective recognition systems using traditional CIL approaches. To address this limitation, this paper focuses on Few-Shot Class-Incremental Learning (FSCIL), a more challenging scenario where only limited data is available. In FSCIL \cite{fcil_surv,fcil_fact,fcil_vt_cap,fcil_savc,fcil_vt_icec,fcil_teen,fcil_vt_prompt,fcil_clom}, the initial model is trained with sufficient samples in the base task, and new classes with only limited data are then incrementally introduced in subsequent tasks. This process simulates the human ability to continually learn new concepts from few samples, aiming to enhance generalization to downstream new classes while maintaining performance on pre-defined base classes \cite{fcil_clom}.

In addition to catastrophic forgetting, learning from few-shot samples in incremental tasks can also lead to overfitting, thereby degrading FSCIL performance. To alleviate this, forward-compatible learning, which prospectively prepares for future tasks during the base task training, has attracted increasing attention. FACT \cite{fcil_fact} first introduced the concept of forward compatibility in FSCIL by assigning virtual prototypes to squeeze the embeddings of known classes, thus reserving embedding space for future classes. Inspired by the dilemma in which positive-margin loss functions, although improving performance on pre-defined data, can hinder generalization to downstream tasks, Zou et al. \cite{fcil_clom} proposed a margin-based FSCIL method, CLOM. This method adds an auxiliary classifier with an intermediate mapping layer on top of the original embedding output, and applies negative margin optimization to the standard classifier and positive margin optimization to the auxiliary classifier during base task learning. By concatenating the logits from both classifiers for final prediction, CLOM enhances generalization to new classes while maintaining strong base class performance. More recently, Liu et al. \cite{asp} proposed ASP, an FSCIL method that incorporates Parameter-Efficient Fine-Tuning (PEFT) on pre-trained Vision Transformer (ViT) \cite{vit}, inspired by the success of PEFT-based methods in traditional CIL research. ASP utilizes task-invariant prompts to encode shared knowledge and task-adaptive prompts to capture task-specific information. This design enables effective knowledge transfer from base to new classes while mitigating forgetting, leading to substantial performance improvements in FSCIL. Nevertheless, these methods still exhibit limited generalization to new classes, hindering their ability to achieve more balanced performance across both base and new classes, as demonstrated by our experimental results in Section~\ref{sec_exp_ana}. This motivates our continued exploration in FSCIL, particularly focusing on enhancing forward compatibility within the PEFT paradigm through margin penalty optimization.

\begin{figure}[tbp]
\centering
\captionsetup[subfigure]{font=footnotesize, labelfont={footnotesize}} 
\subfloat[\scriptsize Negative Correlation Between the Accuracy of Base and New Classes]{\includegraphics[width=0.48\textwidth]{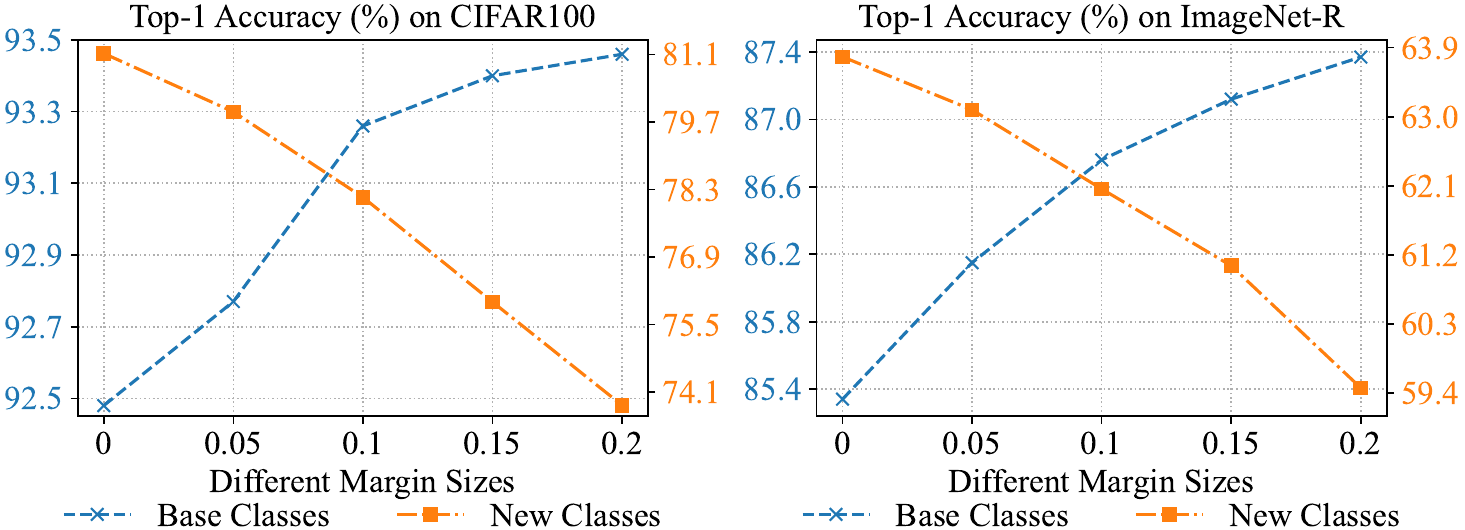}\label{f_neg_trend}}
\\ \vspace{-0.5em}
\subfloat[\scriptsize Can Model Merging Combine Complementary Capabilities?]{\includegraphics[width=0.48\textwidth]{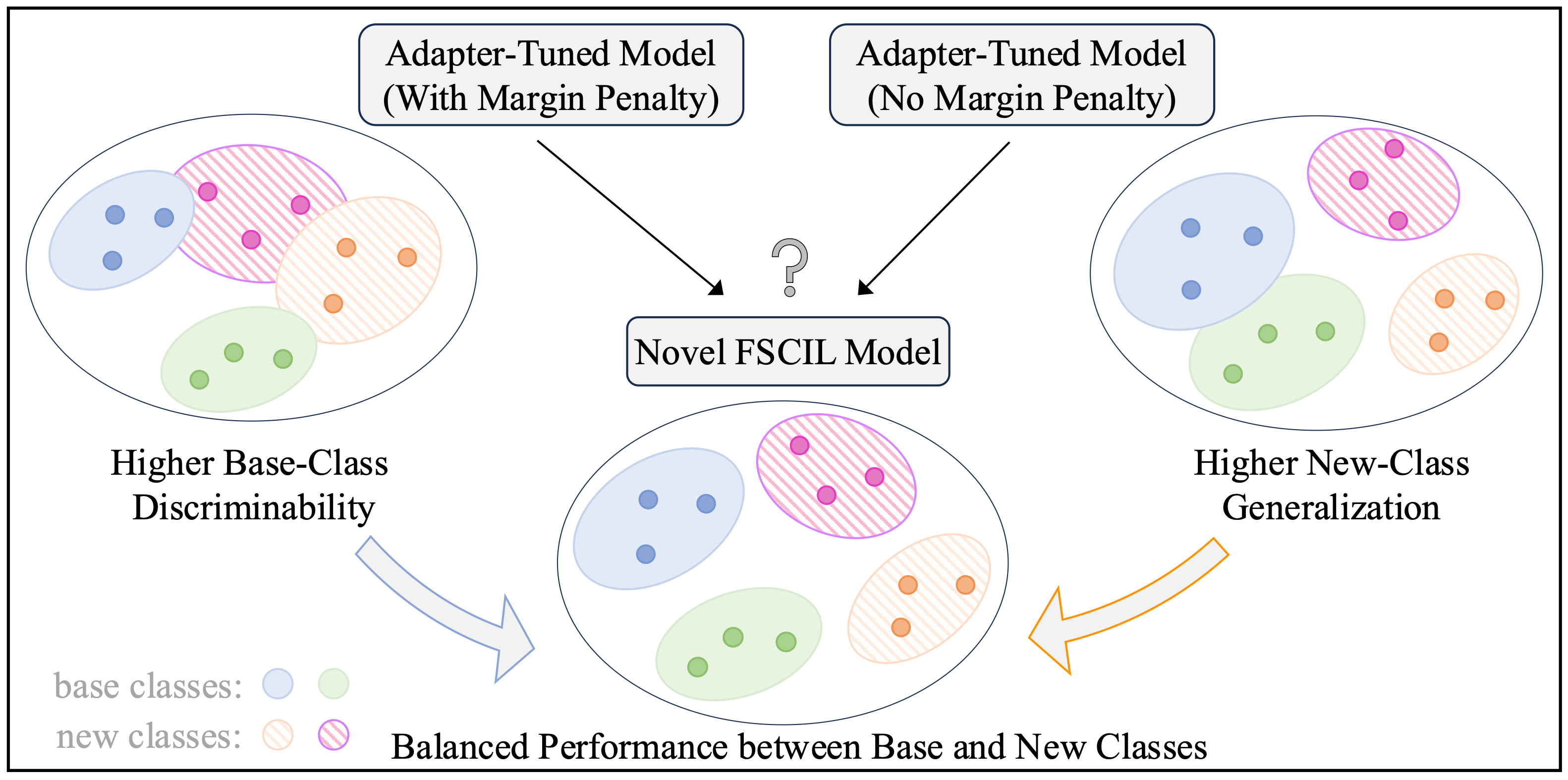}\label{f_merge_q}}\vspace{-0.3em}
\caption{\small (a) Illustration of the trade-off between base-class discriminability and new-class generalization in FSCIL as the margin penalty increases. (b) Motivation for merging complementary models to achieve a more balanced performance across base and new classes.}\vspace{-12pt}
\label{fig:chart}
\end{figure}

To investigate the impact of margin penalties on PEFT-based FSCIL methods, we establish a baseline by fine-tuning a pre-trained ViT model with low-rank adapters in the incremental learning setting. We then conduct two groups of experiments on the widely-used datasets CIFAR100 \cite{cifar} and ImageNet-R \cite{imr} by imposing different margin penalties in the classification loss. As shown in Figure \ref{f_neg_trend}, increasing the margin size improves base class accuracy, but comes at the cost of degraded performance on few-shot new classes. This observation highlights a clear trade-off between base class discriminability and generalization to future new classes, consistent with the findings reported in CLOM under the training-from-scratch paradigm using convolutional neural networks (CNNs). Given that applying different classification objectives to the same ViT backbone can cause mutual interference, especially when using low-rank adapters with limited parameters, we pose the following question: Can we combine two independently trained models—one optimized for better base class discriminability and the other have better generalization to future new classes—into a single FSCIL model, as illustrated in Figure \ref{f_merge_q}, thereby achieving balanced performance by leveraging their complementary strengths?

To realize this idea, we propose the Margin-aware Intra-task Adapter Merging (MIAM) mechanism. Specifically, during base task learning, we introduce two identical sets of low-rank adapters, each independently optimized with distinct objectives to fine-tune the pre-trained ViT. The first set is trained using a classification loss incorporating a large margin penalty to enhance the discriminability among base classes, while the second set is optimized using a classification loss without margin constraints to improve generalization to future new classes. In each transformer layer, each adapter set defines two pairs of low-rank adapters: one pair fine-tuning the key projection matrix, and the other fine-tuning the value projection matrix in the self-attention module. After training, we treat the product of each pair of low-rank adapters within a set as a basic update block, then pairwise merge corresponding blocks between the two sets using adaptively weights derived from their Fisher Information matrices. These weights reflect the blocks' relative importance. The merged adapters are integrated into the original ViT, yielding the base task model. Unlike typical inter-task merging in multi-task learning \cite{local_merge,emr_merge,arith}, MIAM performs intra-task merging by training two adapter sets with complementary abilities on the same base task dataset. Additionally, the low-rank adapters modify only a small portion of the ViT, leaving the backbone largely frozen. This design minimizes interference between the adapter sets, enabling the merged adapter set to effectively combine their strengths, as evidenced by the experimental results and detailed analysis in Section \ref{sec_exp}.

During the incremental learning stage, we follow the mainstream practice of freezing the feature extractor to maintain model stability, while enabling plasticity by constructing classifiers for new classes based on their corresponding prototypes (i.e., class means). However, since the original samples from previous tasks are unavailable and new-class samples are limited, the constructed prototypes may deviate from the true data distribution. This often leads to ambiguous decision boundaries, especially when the class distributions are complex. In the training-from-scratch FSCIL paradigm with CNNs, TEEN \cite{fcil_teen} partially addressed this issue by generating new class classifiers through a fusion of new class prototypes and a weighted combination of base class prototypes. Under the PEFT paradigm, our proposed MIAM mechanism significantly mitigates decision boundary ambiguity through improved generalization to new classes. Nevertheless, this challenge remains to some extent.
To further reduce this ambiguity, we introduce a Margin Penalty-based Classifier Calibration (MPCC) strategy. MPCC employs a classification loss with a large margin penalty to fine-tune classifiers on feature embeddings encompassing all seen classes, thereby enhancing inter-class discrimination and improving overall FSCIL performance.

Finally, we integrate the two key components described above into our baseline model, forming the proposed efficient few-shot class-incremental learning method with Sculpting Margin Penalty (SMP). Our main contributions are summarized as follows:
\begin{itemize}
\item We propose a novel FSCIL method SMP that follows the parameter-efficient fine-tuning paradigm by integrating low-rank adapters. SMP strategically incorporates margin penalty during different learning tasks to achieve efficient incremental learning from limited samples.

\item Within SMP, we introduce a Margin-aware Intra-task Adapter Merging (MIAM) mechanism. MIAM adaptively merges two sets of adapters, one optimized for enhancing base class discriminability and the other for improving generalization to future new classes, thereby improving the forward compatibility of the base task model.

\item Additionally, we propose a Margin Penalty-based Classifier Calibration (MPCC) strategy to alleviate decision boundary ambiguity during incremental tasks, further improving FSCIL performance.

\item Extensive experiments on CIFAR100, ImageNet-R, and CUB200 demonstrate that SMP achieves state-of-the-art performance in FSCIL while maintaining a better balance between base and new classes.
\end{itemize}

\section{Related Work}

\subsection{Few-Shot Class-Incremental Learning}
Few-Shot Class-Incremental Learning focuses on addressing the challenge of learning new classes from a limited number of samples while mitigating catastrophic forgetting of previously learned classes. Early FSCIL research primarily adopted the training-from-scratch paradigm with convolutional neural networks. Meta-learning methods \cite{meta_p1,meta_p2,meta_p4,meta_m1} leverage prototype learning or meta processes to reduce dependence on new task data. Dynamic network structure-based methods \cite{dynamic_1,dynamic_cec,dynamic_3} adjust model architecture to enhance adaptability to incremental tasks. Replay-based methods \cite{replay_d1,replay_d2,replay_g1,replay_g2} mitigate forgetting by revisiting previous task information, either the stored subset of original samples \cite{replay_d1,replay_d2}, or pseudo data through generative models \cite{replay_g1,replay_g2}. Feature space-based methods \cite{feat_1,feat_2,feat_limit} project new class data into the subspace composed of old class features, thereby enabling better adaptation to new classes. Building on these directions, forward-compatible learning has emerged as a promising approach by prospectively prepares for future tasks during base task training. Zhou et al. \cite{fcil_fact} first introduced this concept in FSCIL with the Forward-Compatible Training (FACT) strategy, which reserves embedding space for future classes by assigning virtual prototypes to compress known classes. Song et al. \cite{fcil_savc} extend this approach by pre-allocating virtual placeholders for future classes and applying contrastive learning to enhance both base class separation and new class generalization. Additionally, CLOM \cite{fcil_clom} incorporates margin penalty-based optimization to enhance the transferability of the base model to future classes while preserving accuracy on known classes. Recently, as research on foundational visual models deepens, the PEFT paradigm using pre-trained ViTs \cite{vit} has made significant advancements over the training-from-scratch paradigm in CIL, offering new insights for FSCIL research.

PEFT-based methods typically fine-tune pre-trained models by introducing learnable prompts or low-rank adapters \cite{peft_survey}. L2P \cite{l2p} defined a learnable prompt pool and selected different combinations to guide incremental task learning. DualPrompt \cite{dual} extended this idea by introducing two types of complementary prompts to the pre-trained backbone, including general (task-invariant) prompts and expert (task-specific) prompts. CODA \cite{coda} replaced the prompt pool with weighted learnable components, enabling end-to-end optimization and improved adaptability to future tasks. In addition, InfLoRA \cite{inflora} and LoRA$^-$DRS \cite{lora_sub} add low-rank adapters to the self-attention module of pre-trained ViT for PEFT. Despite these advances, these methods rely on the assumption that sufficient training data is available for each incremental task, leading to performance degradation when new classes have fewer samples. To address this limitation, Liu et al. \cite{asp} proposed the prompt-based FSCIL method ASP, which uses task-invariant prompts to encode shared knowledge and task-adaptive prompts to capture task-specific information. This approach facilitates effective knowledge transfer from base to new classes while mitigating forgetting, resulting in significant performance improvements in FSCIL. Inspired by these existing works, we are committed to continuing to improve the forward compatibility of the FSCIL model under the PEFT paradigm.

\subsection{Margin Penalty-based Classification Optimization}

Margin penalty-based optimization was first introduced for face recognition tasks with convolutional neural networks to enhance the discriminability of the recognition model. Liu et al. \cite{sphere} introduced the angular softmax loss function, SphereFace, which incorporates multiplicative angular margins to enforce intra-class compactness and inter-class separability. However, the objective function in SphereFace makes the target logit curve very precipitous and thus hinders model convergence \cite{cosfact}. Subsequent works \cite{cosfact,add_margin} reformulated the softmax loss as a cosine loss and introduced a cosine margin to maximize the decision boundary in the angular space, leading to better performance than SphereFace. Wang et al. \cite{Arcface} further proposed computing the angle between features and class weights using the inverse cosine function, adding an additive angular margin to the target angle, and then applying the cosine function to obtain the final logits, thereby improving both discriminability and training stability. Liu et al. \cite{neg_margin} summarized that, while introducing negative margins in CNNs may slightly reduce feature separability on the training set, it helps prevent new class samples from being misclustered into multiple peaks, thereby enhancing the generalization to new classes. 

In the CNN-based FSCIL research, Zou et al. \cite{fcil_clom} found that individually applying margin constraints during base class training often fails to balance base class performance and new class generalization. They attributed this to class-level overfitting, where margin-based pattern learning is too easily satisfied. To mitigate this issue, they proposed a margin-based FSCIL method, CLOM, which introduces an auxiliary classifier with an intermediate mapping layer on top of the original embedding output. During base task training, CLOM applies negative margin optimization to the standard classifier and positive margin optimization to the auxiliary classifier. The logits from both classifiers are then concatenated for final prediction, thus enhancing generalization to new classes while preserving strong base class performance. Inspired by these existing works, we are dedicated to further enhancing the forward compatibility of the FSCIL model within the PEFT paradigm.

\section{Methodology}

\begin{figure*}[htb]
\centering
\includegraphics[width=\textwidth]{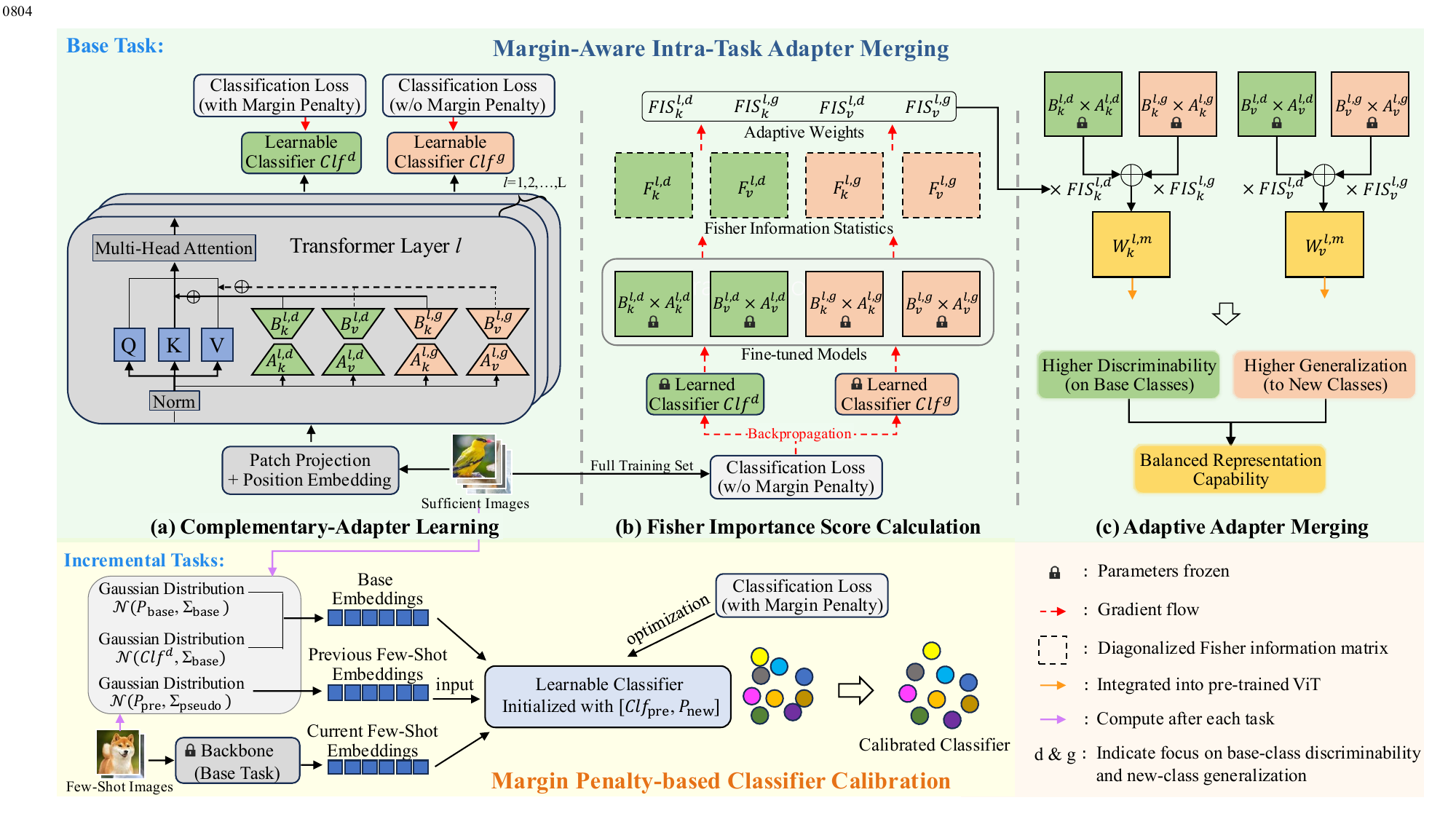}
\caption{\small Overall framework of the proposed SMP for FSCIL. In the base task, we introduce a margin-aware intra-task adapter merging mechanism that adaptively merges two sets of adapters: one optimized for base class discriminability and the other for generalization to future classes, thereby enhancing the model’s forward compatibility. For incremental tasks, we propose a margin penalty-based classifier calibration strategy to alleviate decision boundary ambiguity.}
\label{f_line}
\end{figure*}

\subsection{Problem Definition and Overall Framework} \label{sec:method_1}
Few-Shot Class-Incremental Learning (FSCIL) aims to enable a model to continuously incorporate new classes with only a few labeled samples per class, while maintaining robust recognition capabilities for previously learned classes. The learning process involves a sequence of tasks (sessions), where each task $t$ introduces a mutually exclusive set of classes, $\mathcal{C}_t$. The cumulative class set up to the final task $T$ is defined as $\mathcal{C} = \cup_{t=0}^{T} \mathcal{C}_t$, with $\mathcal{C}_i \cap \mathcal{C}_j = \varnothing$ for all $ i \ne j$. For each task $t$, the training dataset is denoted as $\mathcal{D}_t^{\text{train}} = \left\{(x_i, y_i)\right\}_{i=1}^{N_t}$, where $x_i$ is an input image, $y_i \in \mathcal{C}_t$ is its corresponding label, and $N_t$ is the total number of samples. The initial task ($t = 0$), known as the base task, provides sufficient labeled data for model initialization. Following this, each incremental task ($t > 0$) introduces $N$ new classes, each with only $K$ samples, adhering to the standard $N$-way $K$-shot setting. When evaluating the model at task $t$, the test set $\mathcal{D}_t^{\text{test}}$ includes all classes learned so far. 

As shown in Figure \ref{f_line}, our method strategically integrates margin penalty-based optimization at different stages of FSCIL under the parameter-efficient fine-tuning (PEFT) paradigm, enabling efficient incremental learning of new classes with a limited number of samples. During base task training, we first learn two complementary sets of low-rank adapters using distinct classification objectives: one with a margin penalty to enhance base-class discriminability and the other without margin constraints to encourage generalization to new classes. These adapters are then adaptively merged based on their Fisher importance scores, effectively preserving their complementary strengths. The merged adapters are integrated into the pre-trained ViT to form the base model with improved forward compatibility. In incremental tasks, we first construct feature embeddings for all seen classes, either from real samples or synthesized using Gaussian distributions. Then, we introduce a classification loss with a large margin penalty to fine-tune the classifiers, enhancing inter-class separation and reducing decision boundary ambiguity.

In the following sections, we explain our method from three key parts: parameter-efficient fine-tuning with low-rank adapters for FSCIL, the margin-aware intra-task adapter merging mechanism, and the margin penalty-based classifier calibration strategy.

\subsection{Parameter-Efficient Fine-Tuning with Low-Rank Adapters for FSCIL}
We construct our baseline model within the PEFT paradigm by integrating a set of low-rank adapters into a pre-trained ViT. The ViT backbone consists of a patch embedding layer followed by a stack of transformer layers. Each transformer layer contains a Multi-Head Self-Attention (MHSA) module and a Feed-Forward Network (FFN). In the MHSA module, the input feature matrix $X \in \mathbb{R}^{N \times d}$ (where $N$ is the number of patches and $d$ is the embedding dimension) is projected into query, key, and value matrices using projection weights $W_q$, $W_k$, and $W_v$:
\begin{equation}
Q = XW_q, \quad K = XW_k, \quad V = XW_v.
\end{equation}
The attention output is then computed as:
\begin{equation}
\text{Attention}(Q, K, V) = \text{Softmax} \left( \frac{QK^\top}{\sqrt{d_k}} \right) V,
\end{equation}
\noindent where $d_k$ denotes the dimension of the key vectors in each attention head.
To adapt the pre-trained ViT to the base task in the FSCIL setting, we inject low-rank adapters to fine-tune the key and value projection matrices of each MHSA module, following prior work~\cite{inflora,lora_sub}. Specifically, for each of the frozen original projection matrices $W_k^{(l,0)}$ and $W_v^{(l,0)}$ in transformer layer $l \in \{1, 2, \ldots, L\}$, we introduce a pair of trainable low-rank adapters, $A \in \mathbb{R}^{r \times d}$ and $B \in \mathbb{R}^{d \times r}$. Here, $r \ll d$ significantly reduces the number of learnable parameters. The updated projection matrices are computed as:
\begin{equation}
\begin{aligned}
W_k^{(l)} &= W_k^{(l,0)} + B_k^{(l)} A_k^{(l)} = W_k^{(l,0)} + \Delta W_k^{(l)}, \\
W_v^{(l)} &= W_v^{(l,0)} + B_v^{(l)} A_v^{(l)} = W_v^{(l,0)} + \Delta W_v^{(l)}.
\end{aligned}
\end{equation}
\noindent These basic update blocks (i.e., the product of each pair of low-rank adapters), $\Delta W_k^{(l)}$ and $\Delta W_v^{(l)}$, can be seamlessly added to the frozen projection weights, facilitating efficient task-specific adaptation of the transformer layer. The complete set of learnable low-rank parameters across all layers is defined as:
\begin{equation}
\mathcal{R} = \left\{ A_k^{(l)}, B_k^{(l)}, A_v^{(l)}, B_v^{(l)} \right\},\quad l \in \{1,2,\dots,L\}.
\end{equation}
For the base task, the low-rank adapter set $\mathcal{R}$ and the classifier for the pre-defined base classes are jointly trained using sufficient base class data. Once training is complete, the classifier is replaced with class prototypes computed from the training samples, resulting in the final base-task model.

During the incremental learning stage, we freeze the ViT backbone equipped with the trained adapter set $\mathcal{R}$ and expand the classifier to include new classes by computing their class prototypes. The frozen backbone is denoted as $F(\cdot)$, and the prototype of each class is computed as the mean feature embeddings extracted from its training samples, as follows:
\begin{equation}
\mu_c = \frac{1}{|S_c|} \sum_{x \in S_c} F(x),
\label{eq_proto}
\end{equation}
where $S_c$ is the set of training samples for class $c$.
The two components introduced in the following sections further improve the balance between model stability and plasticity in the FSCIL setting.

\subsection{Margin-Aware Intra-Task Adapter Merging}
In the base task training, we employ the Margin-aware Intra-task Adapter Merging (MIAM) mechanism to enhance forward compatibility. 

\noindent\textbf{Complementary-Adapter Learning.} As shown in Figure~\ref{f_line}\textcolor{blue}{(a)}, we define two sets of low-rank adapters:
\begin{equation}
\begin{aligned}
    \mathcal{R}^{d} = \{A_k^{(l,d)}, B_k^{(l,d)}, A_v^{(l,d)}, B_v^{(l,d)}\},l \in \{1,2,\dots,L\}, \\
\mathcal{R}^{g} = \{A_k^{(l,g)}, B_k^{(l,g)}, A_v^{(l,g)}, B_v^{(l,g)}\},l \in \{1,2,\dots,L\}. 
\end{aligned}
\end{equation}
Each adapter set, combined with the pre-trained ViT backbone and a learnable classifier, constitutes an independent classification model. Given a training sample $(x_i, y_i)$, we extract the feature embedding from the backbone:
\begin{equation}
    f_i = F(x_i) \in \mathbb{R}^d.
\end{equation}
Let the weight matrix of the classifier be $ W = [w_1, w_2, \dots, w_{\left |\mathcal{C}_0\right |}] \in \mathbb{R}^{d \times {\left |\mathcal{C}_0\right |}}$, the classification logit for each class $j \in \mathcal{C}_0$ is computed using scaled cosine similarity:
\begin{equation}
    s_{j}=s \cdot \cos \left(\theta_{j}\right)=s \cdot \frac{w_{j}^{\top} f_{i}}{\left\|w_j\right\| \left\|f_{i}\right\|},
\end{equation}
where $s$ is a scaling factor that controls the scale of the angular distance between the classifier vector and the feature embedding. To promote complementary capabilities, the two classification models are optimized with different objectives. The adapter set $\mathcal{R}^{d}$ and its classifier $Clf^d$ are trained to enhance base class discriminability using a classification loss with a large margin penalty. This margin penalty-based classification loss, also referred to as the discriminative classification loss, is defined as:
\begin{equation}
\mathcal{L}^{d} = -\log \frac{e^{s \cdot \left( \cos(\theta_{y_{i}}) - m \right)}}{e^{s \cdot \left( \cos(\theta_{y_{i}}) - m \right)} + \sum_{j=1, j \neq y_{i}}^{|\mathcal{C}_0|} e^{s \cdot \cos(\theta_{j})}},
\end{equation}
\noindent where $m$ is a margin size hyperparameter that controls the degree of angular separation between class vectors, thereby enhancing class discriminability.
 
Meanwhile, the adapter set $\mathcal{R}^{g}$, together with the learnable classifier $Clf^{g}$, is trained to improve generalization to future new classes using a classification loss without margin constraints. This plain classification loss, referred to as the generalization classification loss, is defined as:
\begin{equation}
\mathcal{L}^{g}=-\log \frac{e^{s\cos \left(\theta_{y_{i}}\right)}}{\sum_{j=1}^{{|\mathcal{C}_0|}} e^{s \cos \left(\theta_{j}\right)}}.
\end{equation}
\noindent It is worth noting that, unlike prior CIL methods \cite{inflora,lora_sub} which treat the angular distance scale $s$ as a learnable parameter, we follow CLOM \cite{fcil_clom} and fix $s$ to a larger value of 16 in our MIAM mechanism. This setting amplifies the separation between class logits, enabling the generalization-oriented loss $\mathcal{L}^{g}$ to achieve better generalization to new classes, while the discriminative loss $\mathcal{L}^{d}$ benefits from the margin penalty. For the margin size $m$, we set it to 0.2 and demonstrate its robustness through the experiments in Figure \ref{f_m_size}.

\noindent \textbf{Fisher Importance Score Calculation.} 
As shown in Figure~\ref{f_line}\textcolor{blue}{(b)}, after training the two complementary classification models, we estimate the relative importance of each update block (i.e., the product of each pair of low-rank adapters) in the two sets using their Fisher Information statistics under the generalization loss $\mathcal{L}^{g}$. The Fisher Information statistic is computed over the entire base task training dataset. Following EWC~\cite{ewc}, we approximate the Fisher Information statistic using the diagonal of the standard Fisher Information matrix. Using $\Delta W^{(l)} = B^{(l)}A^{(l)}$ to represent any of the basic update blocks applied to the key and value projection matrices, the Fisher Information statistics for the discriminative and generalizable classification models are estimated as follows:
\begin{equation}
\begin{aligned}
F^{(l,d)} &=\mathbb{E}_{x \sim \mathcal{D}_{\text {0 }}^{\text{train}}}\left[\left(\frac{\partial \mathcal{L}^{g}}{\partial \Delta W^{(l,d)}}\right)^{2}\right],\\
F^{(l,g)} &=\mathbb{E}_{x \sim \mathcal{D}_{\text {0 }}^{\text{train}}}\left[\left(\frac{\partial \mathcal{L}^{g}}{\partial \Delta W^{(l,g)}}\right)^{2}\right].
\end{aligned}
\end{equation}
We then compute the Frobenius norms of the Fisher Information statistics and normalize them to obtain the relative importance scores of the update blocks in the two models:
\begin{equation}
\begin{aligned}
\begin{aligned}
FIS^{(l,d)} = \frac{\left| F^{(l,d)} \right|_F}{\left| F^{(l,d)} \right|_F + \left| F^{(l,g)} \right|_F},\\
 FIS^{(l,g)} = \frac{\left| F^{(l,g)} \right|_F}{\left| F^{(l,d)} \right|_F + \left| F^{(l,g)} \right|_F}.
\end{aligned}
\end{aligned}
\end{equation}
\noindent \textbf{Adaptive Adapter Merging.}
As illustrated in Figure~\ref{f_line}\textcolor{blue}{(c)}, the update blocks from both discriminative and generalizable models are merged in each transformer layer based on their importance scores, as follows:
\begin{equation}
\begin{aligned}
\Delta W^{(l,m)}_{k} &= FIS^{(l,d)}_{k} \cdot \Delta W^{(l,d)}_{k} + FIS^{(l,g)}_{k} \cdot \Delta W^{(l,g)}_{k},\\
\Delta W^{(l,m)}_{v} &= FIS^{(l,d)}_{v} \cdot \Delta W^{(l,d)}_{v} + FIS^{(l,g)}_{v} \cdot \Delta W^{(l,g)}_{v}.
\end{aligned}
\end{equation}
Applying this merging strategy across all layers yields the final set of adapters:
\begin{equation}
\mathcal{R}^{m} = \{ \Delta W^{(l,m)}_{k}, \Delta W^{(l,m)}_{v}\},l \in \{1,2,\dots,L\}.
\end{equation}
These merged adapters are then integrated into the frozen ViT backbone, forming the backbone for the base task. Together with the base class prototypes computed through Equation~\ref{eq_proto}, this forms the final model for the base task. This design enables the model to maintain strong discriminability for base classes and superior generalization for future new classes. In Section \ref{sec_exp}, we validate the effectiveness of MIAM and analyze the reasons why it successfully balances these complementary capabilities.

\subsection{Margin Penalty-based Classifier Calibration}
As shown in the lower part of Figure~\ref{f_line}, for each incremental task, we first construct a comprehensive embedding training set $F_{\text{input}}$, which includes all classes encountered up to the current task.

Inspired by works \cite{hide,ssiat}, we prepare feature embeddings for previous classes using Gaussian distribution sampling. For each base class $co \in \mathcal{C}_0$ , we generate pseudo-features by evenly sampling from two multivariate Gaussian distributions: one centered at the statistical prototype $\mathcal{N}(P_{\text{base}}^{co}, \Sigma_{\text{base}}^{co})$, and the other centered at the classifier vectors $\mathcal{N}(Clf^{d}_{co}, \Sigma_{\text{base}}^{co})$. Here, $Clf^{d}_{co}$ denotes the slice of the learned weights in $Clf^{d}$ corresponding to class $co$. The prototype $P_{\text{base}}^{co}$ and the covariance matrix $\Sigma_{\text{base}}^{co}$ for class $co$ are estimated after base task training, as follows:
\begin{equation}
\begin{aligned}
P_{\text{base}}^{co} &= \frac{1}{N_{co}} \sum_{i=1}^{N_{co}} f_i, \\
\Sigma_{\text{base}}^{co} &= \frac{1}{N_{co}} \sum_{i=1}^{N_{co}} \left( f_i - P_{\text{base}}^{co} \right) \left( f_i - P_{\text{base}}^{co} \right)^\top,
\end{aligned}
\end{equation}
where $N_{co}$ is the total number of samples in class $co$, and $f_i$ denotes the feature embedding of the $i$-th sample.
For each few-shot class from previous incremental tasks, denoted by $cp \in C_{\text{pre}}$ where $C_{\text{pre}} = {\textstyle \bigcup_{i=1}^{t-1}} C_i$, we approximate its covariance using that of its most semantically similar base class $c'$, since the covariance matrix estimated from limited samples is often unstable and noisy. This approximation is defined as:
\begin{equation}
\Sigma_{\text{pseudo}}^{cp} = \Sigma_{\text{base}}^{c'},\, \text{where} \, c' = \arg\max_{j \in \mathcal{C}_0} \cos(P_{\text{pre}}^{cp}, P_{\text{base}}^j),
\end{equation}
\noindent with $P_{\text{pre}}^{cp}$ and $P_{\text{base}}^j$ denoting the prototypes of class $cp$ and the base class $j$, respectively. We then sample pseudo-features for class $cp$ from the Gaussian distribution $\mathcal{N}(P_{\text{pre}}^{cp}, \Sigma_{\text{pseudo}}^{cp})$. For each few-shot class in the current incremental task $t$, $ct \in C_t$, we directly extract feature embeddings from real image samples of the class. To construct a balanced embedding training set $F_{\text{input}}$, we ensure that each class has an equal number of samples (256) per training iteration.

Next, we compute the prototypes of the few-shot new classes in the current task $t$ using Equation \ref{eq_proto}, denoted as $P_{\text{new}}$. We then concatenate $P_{\text{new}}$ with the classifier weights of all previously learned classes acquired from task $t-1$, denoted as $Clf_{\text{pre}}$, to construct the initial classifier for the current task $t$, as follows:
\begin{equation}
W_\text{init}=[Clf_{\text{pre}}, P_{\text{new}}].
\end{equation}
This classifier is then optimized using the margin penalty-based classification loss. Let $(\bar{f}_{i},\bar{y}_{i})$ represent each sample in $F_{\text{input}}$, the classification loss is defined as:
\begin{equation}
\mathcal{L}_{\text{calib}} = -\log \frac{e^{s(\cos(\theta{\bar{y}_i}) - m)}}{e^{s(\cos(\theta_{\bar{y}_i}) - m)} + \sum_{j=1,j \ne \bar{y}_i}^{\left | \bigcup_{k=0}^t \mathcal{C}_k \right |} e^{s \cos(\theta_j)}},
\end{equation}
\noindent where $s$ represents the scaling factor and $m$ denotes the margin size, both set to the same values used in the MIAM mechanism. The calibrated classifier weights $W_{\text{cab}}$, obtained after training, are used as the final classifier for inference in the current incremental task.

\section{Experiments and Analysis} \label{sec_exp}
\subsection{Experimental Settings}
\noindent\textbf{Datasets.}
Following FSCIL works \cite{fcil_teen,fcil_clom,asp}, we evaluate the performance of our method on three widely used datasets: CIFAR100 \cite{cifar}, ImageNet-R \cite{imr}, and CUB-200-2011 \cite{cub200}.
CIFAR100 consists of 100 classes with a total of 60,000 images, where each class contains 500 training images and 100 test images. The dataset is split into 60 base classes and 40 incremental classes, with the incremental classes evenly divided across 8 tasks. Each incremental task adopts a 5-way 5-shot setting, introducing 5 new classes with 5 training samples per class.
ImageNet-R comprises 30,000 images across 200 classes. It is divided into 100 base classes and 100 incremental classes, with the incremental classes distributed across 10 tasks. Each incremental task follows a 10-way 5-shot setting, introducing 10 new classes, each with 5 training samples.
CUB-200 is a fine-grained dataset containing 200 classes and 11,788 images. It is similarly divided into 100 base classes and 100 incremental classes, evenly distributed across 10 tasks. Each incremental task adopts the same 10-way 5-shot setting as used in ImageNet-R.

\noindent\textbf{Implementation Details.}
Following previous works \cite{l2p,dual,coda,asp}, we utilize ViT-B/16-1K \cite{vit} (pre-trained on ImageNet-1K) as the initial backbone to construct our low-rank adapter-based FSCIL model. For base task training, we employ the SGD optimizer for 20 epochs across all datasets, with input images resized to 224×224. The learning rate is set to 0.01, and the batch sizes are 48 for CIFAR100 and 24 for both ImageNet-R and CUB-200. In the incremental learning stage, we freeze the feature extractor (i.e., the ViT-B backbone) after integrating the low-rank adapters, and train only the classifier with a learning rate of 0.001. The rank of the adapter pairs is set to 10, following the settings used in InfLoRA \cite{inflora} and LoRA-DRS \cite{lora_sub}. All experiments are conducted with three random seeds, and average results are reported.

We compare our method with widely used CIL methods, including iCaRL \cite{icarl}, Foster \cite{foster}, L2P \cite{l2p}, DualP \cite{dual}, and CodaP \cite{coda}, as well as state-of-the-art FSCIL methods such as CEC \cite{dynamic_cec}, FACT \cite{fcil_fact}, TEEN \cite{fcil_teen}, CLOM \cite{fcil_clom}, ADBS \cite{adbs}, and ASP \cite{asp}. L2P, DualP, CodaP, and ASP adopt learnable prompts to enable efficient fine-tuning, while L2P+, DualP+, and CodaP+ enhance their respective methods by replacing the original classifier with class prototypes. The results for iCaRL, Foster, CEC, FACT, TEEN, L2P, DualP, CodaP, L2P+, DualP+, CodaP+, and ASP are directly cited from \cite{asp}. For CLOM and ADBS, we integrate their core innovations into our built baseline model, ensuring consistency in backbone and experimental setup for a fair comparison with our method. Specifically, for CLOM, we add an auxiliary classifier with an intermediate mapping layer on top of the embedding output, following the original paper. The dimension of this layer is set to 512 on all three datasets. For ADBS, we scale its classification logits by a factor of $s = 16$ and train for 20 epochs during the incremental learning stage to yield the best performance.

\noindent\textbf{Evaluation Metrics.}
Consistent with works \cite{first,fcil_fact,asp}, we employ Top-1 accuracy ($A_t$), final accuracy (i.e., the Top-1 accuracy on the last task, denoted as $A_{T}$ or $A_{final}$), average accuracy ($A_{avg}$), and harmonic accuracy ($\text{HAcc}$) as our evaluation metrics. $A_t$ denotes the average accuracy of the model after learning task $t$, evaluated on all classes observed up to task $t$. $A_{avg}$, defined as $ A_{avg}={\textstyle \sum_{0}^{T}}A_t/(T+1) $, measures the overall performance across all incremental tasks. $\text{HAcc}$ \cite{hacc}, calculated as $\text{HAcc} = \frac{2 \times A_o \times A_n}{A_o + A_n}$, reflects the balanced performance between base and new classes on the final task. Here, $A_o$ represents the accuracy of base classes introduced in task $0$, and $A_n$ is the accuracy of all incrementally added new classes (for $t > 0$).

\begin{table*}[]
\centering
\caption{\small Detailed Top-1 accuracy $A_t$ in each incremental task, average accuracy $A_{avg}$  and Harmonic Accuracy (HAcc) on CIFAR100 dataset.}
\footnotesize
\resizebox{0.75\linewidth}{!}{
\setlength\tabcolsep{2.5mm}
\begin{tabular}{llcccccccccc}
\toprule
\multirow{2}{*}{Method} & \multicolumn{9}{c}{Accuracy $A_t$ in each task   (\%)}            & \multirow{2}{*}{$A_{avg}$} &  \multirow{2}{*}{HAcc} \\ \cline{2-10}
                        & 0             & 1    & 2    & 3    & 4    & 5    & 6    & 7    & 8    &                        &                        \\ \midrule
iCaRL \cite{icarl}      & 94.2          & 88.9 & 84.7 & 80.0 & 74.9 & 75.6 & 71.8 & 68.2 & 67.1 & 78.4                   & 57.5                   \\
Foster \cite{foster}    & 94.2          & 88.3 & 81.6 & 77.0 & 72.8 & 67.9 & 64.4 & 60.9 & 58.3 & 73.9                   & 11.0                   \\ \midrule
CEC \cite{dynamic_cec}  & 91.6          & 88.1 & 85.3 & 81.7 & 80.2 & 78.0 & 76.5 & 74.8 & 72.6 & 81.0                   & 64.1                   \\
FACT \cite{fcil_fact}   & 91.0          & 87.2 & 83.5 & 79.7 & 77.2 & 74.8 & 73.1 & 71.6 & 69.4 & 78.6                   & 55.5                   \\
TEEN \cite{fcil_teen}   & 92.9          & 90.2 & 88.4 & 86.8 & 86.4 & 86.0 & 85.8 & 85.1 & 84.0 & 87.3                   & 81.2                   \\ \midrule
L2P \cite{l2p}          & 92.2          & 85.2 & 79.2 & 73.8 & 69.2 & 65.1 & 61.4 & 58.1 & 55.2 & 71.1                   & 0.0                    \\
DualP \cite{dual}       & 91.8          & 84.7 & 78.6 & 73.3 & 68.7 & 64.6 & 61.1 & 57.8 & 54.9 & 70.6                   & 0.1                    \\
CodaP \cite{coda}       & 93.4          & 86.2 & 80.1 & 74.7 & 70.1 & 66.0 & 62.3 & 59.0 & 56.0 & 72.0                   & 0.0                    \\
L2P+ \cite{l2p}         & 84.7          & 82.3 & 80.1 & 77.5 & 77.0 & 76.0 & 75.6 & 74.1 & 72.3 & 77.7                   & 68.0                   \\
DualP+ \cite{dual}      & 86.0          & 83.6 & 82.9 & 80.2 & 80.6 & 80.2 & 80.5 & 79.0 & 77.4 & 81.1                   & 75.3                   \\
CodaP+ \cite{coda}      & 86.0          & 83.6 & 81.6 & 79.2 & 79.1 & 78.5 & 78.3 & 77.0 & 75.4 & 79.9                   & 72.2                   \\
CLOM \cite{fcil_clom}   & \textbf{94.8} & 92.3 & 91.3 & 89.4 & 89.1 & 88.2 & 88.0 & 87.3 & 85.9 & 89.6                   & 82.7                   \\
ASP \cite{asp}          & 92.2          & 90.7 & 90.0 & 88.7 & 88.7 & 88.2 & 88.2 & 87.8 & 86.7 & 89.0                   & 85.3                   \\
ADBS \cite{adbs}        & 93.3          & 91.4 & 90.4 & 89.2 & 89.0 & 88.4 & 88.2 & 87.6 & 86.5 & 89.3                   & 84.7                   \\
SMP (Ours)              & 94.5          & \textbf{92.7} & \textbf{92.0} & \textbf{90.7} & \textbf{90.7} & \textbf{90.2} & \textbf{90.1} & \textbf{89.6} & \textbf{88.7} & \textbf{91.0}                   & \textbf{87.1}                   \\ \bottomrule 
\end{tabular}}
\label{main_cifar}
\end{table*}

\begin{table*}[]
\centering
\caption{\small Detailed Top-1 accuracy $A_t$ in each incremental task, average accuracy $A_{avg}$  and Harmonic Accuracy (HAcc) on ImageNet-R dataset.}
\setlength\tabcolsep{2.8mm}
\resizebox{0.9\linewidth}{!}{
\begin{tabular}{llcccccccccccc}
\toprule
\multirow{2}{*}{Method} & \multicolumn{11}{c}{Accuracy $A_{t}$ in each task (\%)}                         & \multirow{2}{*}{$A_{avg}$} & \multirow{2}{*}{HAcc} \\ \cline{2-12}
                        & 0    & 1    & 2    & 3    & 4    & 5    & 6    & 7    & 8    & 9    & 10   &                        &                        \\ \midrule
iCaRL \cite{icarl}      & 80.6 & 69.2 & 59.0 & 52.8 & 49.4 & 45.5 & 42.8 & 42.3 & 40.5 & 40.1 & 39.4 & 51.0                   & 36.1                   \\
Foster \cite{foster}    & 85.8 & 78.8 & 71.8 & 67.4 & 63.1 & 58.5 & 55.9 & 53.8 & 51.5 & 49.3 & 47.0 & 62.1                   & 36.8                   \\ \midrule
CEC \cite{dynamic_cec}  & 79.4 & 71.9 & 69.0 & 64.1 & 60.4 & 58.6 & 56.4 & 53.2 & 52.0 & 50.0 & 48.3 & 60.3                   & 32.6                   \\
FACT \cite{fcil_fact}   & 79.4 & 72.5 & 69.0 & 63.8 & 60.1 & 57.6 & 54.7 & 52.2 & 50.2 & 48.1 & 46.0 & 59.4                   & 22.3                   \\
TEEN \cite{fcil_teen}   & 84.6 & 76.7 & 68.8 & 67.6 & 64.3 & 60.6 & 58.3 & 56.1 & 56.1 & 54.7 & 54.9 & 63.9                   & 45.4                   \\ \midrule
L2P \cite{l2p}          & 80.4 & 73.0 & 67.8 & 62.3 & 58.0 & 55.1 & 52.0 & 48.3 & 45.8 & 42.9 & 40.6 & 56.9                   & 1.0                    \\
DualP \cite{dual}       & 75.6 & 68.5 & 63.8 & 58.7 & 54.7 & 52.2 & 49.4 & 46.0 & 43.8 & 41.1 & 39.0 & 53.9                   & 2.7                    \\
CodaP \cite{coda}       & 82.1 & 74.4 & 69.4 & 64.1 & 59.8 & 57.1 & 53.9 & 50.5 & 48.2 & 45.3 & 43.2 & 58.9                   & 6.5                    \\
L2P+ \cite{l2p}         & 73.9 & 70.9 & 69.3 & 65.9 & 64.0 & 62.6 & 60.1 & 59.5 & 59.0 & 58.2 & 56.8 & 63.7                   & 52.0                   \\
DualP+ \cite{dual}      & 71.5 & 69.0 & 69.0 & 67.4 & 66.6 & 65.9 & 64.1 & 64.0 & 64.0 & 63.4 & 62.5 & 66.1                   & 61.6                   \\
CodaP+ \cite{coda}      & 74.4 & 69.3 & 67.7 & 63.7 & 62.0 & 61.3 & 58.5 & 58.2 & 57.5 & 54.7 & 55.3 & 62.0                   & 54.5                   \\
CLOM \cite{fcil_clom}   & \textbf{88.0} & 84.5 & 83.6 & 80.5 & 79.0 & 78.1 & 76.6 & 75.9 & 75.5 & 74.6 & 73.2 & 79.0                   & 70.2    \\
ASP \cite{asp}          & 83.3 & 80.4 & 79.6 & 77.0 & 75.6 & 74.7 & 73.0 & 72.1 & 71.9 & 70.9 & 69.7 & 75.3                   & 67.0                   \\
ADBS \cite{adbs}        & 85.2 & 83.0 & 82.7 & 80.6 & 79.5 & 78.8 & 77.4 & 76.7 & 76.6 & 76.0 & 75.0 & 79.2                   & 73.6                   \\
SMP (Ours)                & 87.6 & \textbf{85.1} & \textbf{84.6} & \textbf{82.3} & \textbf{81.1} & \textbf{80.4} & \textbf{79.2} & \textbf{78.7} & \textbf{78.5} & \textbf{77.9} & \textbf{76.7} & \textbf{81.1}                   & \textbf{75.1}                   \\ \bottomrule  
\end{tabular}}
\label{main_imr}
\end{table*}

\begin{table*}[]
\centering
\caption{\small Detailed Top-1 accuracy $A_t$ in each incremental task, average accuracy $A_{avg}$  and Harmonic Accuracy (HAcc) on CUB200 dataset.}
\setlength\tabcolsep{2.8mm}
\resizebox{0.9\linewidth}{!}{
\begin{tabular}{llcccccccccccc}
\toprule
\multirow{2}{*}{Method} & \multicolumn{11}{c}{Accuracy $A_{t}$ in each task (\%)}                         & \multirow{2}{*}{\textbf{$A_{avg}$}} & \multirow{2}{*}{HAcc} \\ \cline{2-12}
                        & 0    & 1    & 2    & 3    & 4    & 5    & 6    & 7    & 8    & 9    & 10   &                        &                        \\ \midrule
iCaRL \cite{icarl}      & 92.4 & 82.3 & 71.7 & 62.9 & 64.6 & 62.9 & 60.8 & 60.3 & 58.4 & 55.7 & 58.9 & 66.4                   & 56.4                   \\ 
Foster \cite{foster}    & \textbf{93.0} & 86.4 & 80.4 & 74.5 & 72.9 & 69.3 & 68.5 & 66.0 & 65.9 & 65.0 & 63.4 & 73.2                   & 58.8                   \\ \midrule
CEC \cite{dynamic_cec}  & 84.8 & 82.5 & 81.4 & 78.5 & 79.3 & 77.8 & 77.4 & 77.6 & 77.2 & 76.9 & 76.8 & 79.1                   & 76.2                   \\
FACT \cite{fcil_fact}   & 87.3 & 84.2 & 82.1 & 78.1 & 78.4 & 76.3 & 75.4 & 75.5 & 74.4 & 74.1 & 73.9 & 78.2                   & 72.0                   \\
TEEN \cite{fcil_teen}   & 89.0 & 86.5 & 85.9 & 83.3 & 83.3 & 82.2 & 82.1 & 80.4 & 80.5 & 80.1 & 80.6 & 83.1                   & 80.2                   \\ \midrule
L2P \cite{l2p}          & 87.8 & 81.3 & 74.2 & 68.9 & 63.4 & 59.2 & 55.8 & 51.9 & 49.0 & 46.3 & 44.3 & 62.0                   & 3.3                    \\
DualP \cite{dual}       & 88.9 & 82.6 & 75.3 & 69.6 & 64.5 & 60.4 & 56.8 & 53.0 & 50.0 & 47.5 & 45.5 & 63.1                   & 5.0                    \\
CodaP \cite{coda}       & 89.9 & 83.1 & 75.8 & 70.2 & 65.1 & 61.0 & 57.5 & 53.7 & 50.7 & 48.3 & 46.2 & 63.8                   & 5.7                    \\
L2P+ \cite{l2p}         & 82.4 & 81.2 & 79.0 & 76.8 & 76.2 & 74.7 & 74.1 & 74.1 & 72.7 & 73.0 & 73.6 & 76.2                   & 73.0                   \\
DualP+ \cite{dual}      & 83.5 & 82.2 & 80.9 & 79.5 & 78.6 & 77.0 & 76.3 & 77.0 & 75.7 & 76.1 & 76.5 & 78.5                   & 76.3                   \\
CodaP+ \cite{coda}      & 79.6 & 78.1 & 76.4 & 75.6 & 75.0 & 73.1 & 72.5 & 72.8 & 72.0 & 72.4 & 72.9 & 74.6                   & 72.5                   \\
CLOM \cite{fcil_clom}   & 91.4 & \textbf{90.1} & \textbf{88.6} & \textbf{87.4} & \textbf{87.3} & 85.7 & 85.7 & 86.0 & 85.1 & 85.4 & 85.8 & \textbf{87.1}                   & 85.4 \\
ASP \cite{asp}          & 87.1 & 86.0 & 84.9 & 83.4 & 83.6 & 82.4 & 82.6 & 83.0 & 82.6 & 83.0 & 83.5 & 83.8                   & 83.4                   \\
ADBS \cite{adbs}        & 87.6 & 85.6 & 84.7 & 83.5 & 83.1 & 81.8 & 82.1 & 82.4 & 81.6 & 82.2 & 82.7 & 83.4                   & 82.8                   \\
SMP (Ours)                & 89.8 & 89.2 & 87.8 & 86.9 & 87.2 & \textbf{85.9} & \textbf{86.1} & \textbf{86.5} & \textbf{85.7} & \textbf{86.2} & \textbf{86.6} & \textbf{87.1}   & \textbf{86.4}                   \\ \bottomrule  
\end{tabular}}
\label{main_cub}
\end{table*}

\subsection{Comparison With State-of-the-Art} \label{sec_overall}
\begin{figure*}[htbp]
\centering
\includegraphics[width=\textwidth]{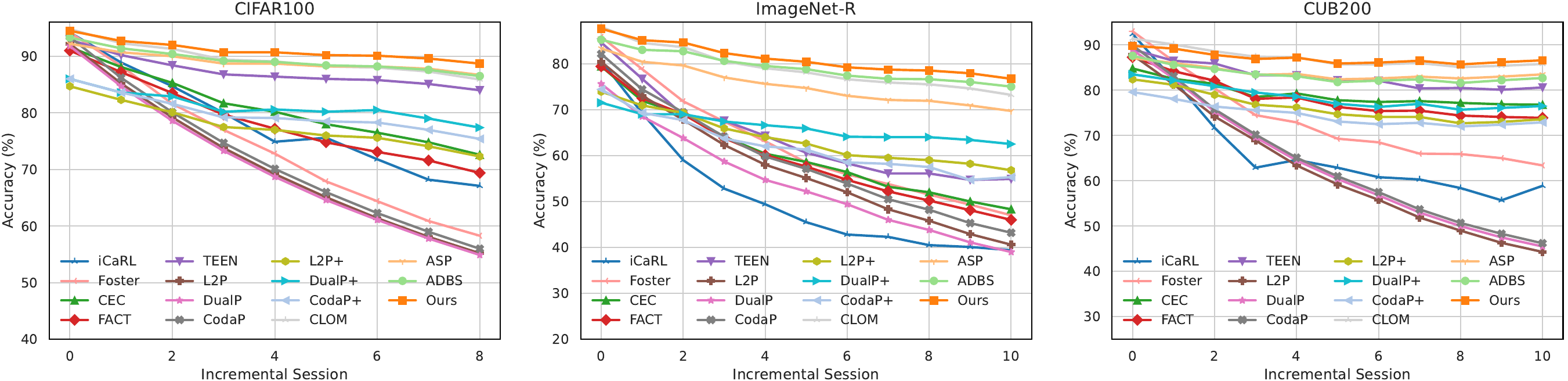}
\caption{\small Detailed Top-1 accuracy $A_t$ of our method and state-of-the-art methods at each incremental task on three benchmark datasets.}
\label{f_main_curvy}
\end{figure*}

This section presents the performance comparison between existing methods and our proposed SMP across the CIFAR100, ImageNet-R, and CUB-200 datasets. Detailed results for these three datasets are provided in Table \ref{main_cifar}, Table \ref{main_imr}, and Table \ref{main_cub}, respectively. Additionally, Figure \ref{f_main_curvy} illustrates the Top-1 accuracy trends across incremental tasks.

The experimental results show that SMP consistently achieves superior performance across all evaluation metrics and datasets. Specifically, in terms of final task accuracy, SMP attains state-of-the-art results on all three datasets, outperforming the second-best method by 2.0\%, 1.7\%, and 0.8\% on CIFAR100, ImageNet-R, and CUB-200, respectively. For average accuracy across all tasks, SMP also records the best performance across all three datasets, surpassing the second-best method by 1.4\% on CIFAR100 and 1.9\% on ImageNet-R. Furthermore, SMP achieves the highest HAcc on all three datasets, exceeding the second-best method by 1.8\%, 1.5\%, and 1.0\%, respectively. This clearly demonstrates SMP’s superior ability to balance performance between base and new classes.

Moreover, as shown in Figure \ref{f_main_curvy}, although SMP does not always achieve the highest accuracy on the base task, it consistently outperforms all other methods on subsequent incremental tasks for both CIFAR100 and ImageNet-R. This highlights its robust resistance to catastrophic forgetting and superior adaptability to new classes. This better performance, also evidenced by higher HAcc results, stems from the integration of the MIAM and MPCC components, where the former promotes forward compatibility by preserving base class performance while improving generalization to future classes, and the latter mitigates decision boundary ambiguity throughout incremental tasks.

\subsection{Performance Analysis} \label{sec_exp_ana}

\begin{table}[htbp]
\centering
\caption{Ablation study of each component in our method. Final accuracy $A_{final}$ and average accuracy $A_{avg}$ are reported on three benchmark datasets.
}
\renewcommand\tabcolsep{2.0mm}
\resizebox{1.0\linewidth}{!}{
\begin{tabular}{cccccccc}
\toprule
 \multirow{2}{*}{MIAM} & \multirow{2}{*}{MPCC}     & \multicolumn{2}{c}{CIFAR100} & \multicolumn{2}{c}{ImageNet-R} & \multicolumn{2}{c}{CUB200} \\
                                                                     &              & $A_{final}$         & $A_{avg}$          & $A_{final}$          & $A_{avg}$           & $A_{final}$        & $A_{avg}$        \\ \midrule
                                                                     &              & 87.4          & 90.0         & 71.7           & 77.7          & 84.2         & 84.8        \\
\checkmark                                                           &              & 88.1          & 90.7         & 75.8           & 80.6          & 85.3         & 86.1        \\
                                                                     & \checkmark   & 88.3          & 90.5         & 72.9           & 78.2          & 85.2         & 85.5        \\
\checkmark                                                           & \checkmark   & \textbf{88.7} & \textbf{91.0} & \textbf{76.7} & \textbf{81.1} & \textbf{86.6} & \textbf{87.1}        \\ \bottomrule
\end{tabular}}
\label{abla}
\end{table}

\noindent \textbf{Ablation Study.}
In Table \ref{abla}, we evaluate the impact of each component of our method on the CIFAR100, ImageNet-R, and CUB-200 datasets. For the baseline method (first row), we utilize a single set of low-rank adapters to fine-tune the pre-trained ViT model, optimizing it with a commonly used classification loss integrated with a learnable parameter to scale logits. During incremental learning tasks, the base task's feature extractor (backbone) is frozen, and the classifier is extended by including prototypes of new classes. In the second, third, and fourth rows, we sequentially introduce the Margin-aware Intra-task Adapter Merging (MIAM) mechanism, the Margin Penalty-based Classifier Calibration (MPCC) strategy, and their combination, respectively.

The experimental results consistently validate the effectiveness of MIAM, MPCC, and their combined use. Specifically, incorporating the MIAM component alone substantially improves performance, with final accuracies on the three datasets increasing by 0.7\%, 4.1\%, and 1.1\%, and average accuracies rising by 0.7\%, 2.9\%, and 1.3\%, respectively. Similarly, introducing MPCC alone into the baseline method also leads to notable performance gains across all datasets. Importantly, the most significant improvements occur when both MIAM and MPCC are employed together, resulting in final accuracy gains of 1.3\%, 5.0\%, and 2.4\%, and average accuracy improvements of 1.0\%, 3.4\%, and 2.3\%, respectively.

\begin{figure*}[htbp]
\centering
\captionsetup[subfigure]{font=footnotesize, labelfont={footnotesize}} 
\hfill
\subfloat[\scriptsize CIFAR100]{\includegraphics[width=0.46\textwidth]{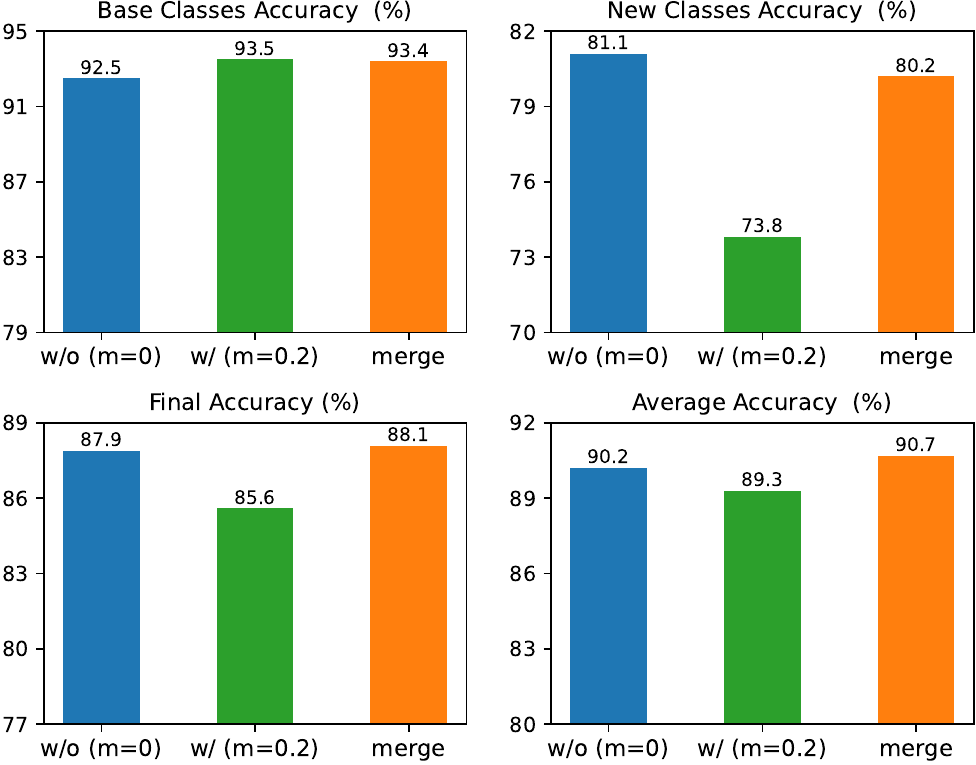}
\label{f_balance_c}}
\hspace{18px}
\subfloat[\scriptsize ImageNet-R]{\includegraphics[width=0.46\textwidth]{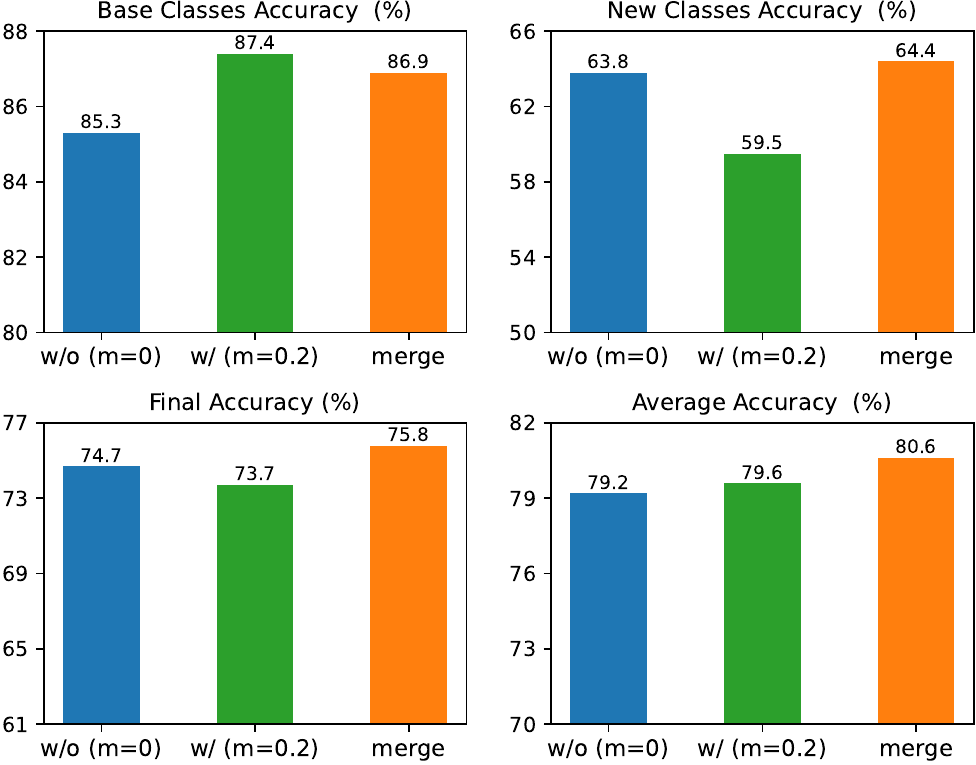}
\label{f_balance_r}}
\hfill
\caption{\small Comparison of base class accuracy, new class accuracy, final accuracy, and average accuracy before and after adapter merging, illustrating the detailed effects of the intra-task adapter merging.}
\label{f_balance}
\end{figure*}

\noindent \textbf{Detailed Effect of Intra-Task Adapter Merging.}
In Figure \ref{f_balance}, we provide a more fine-grained accuracy comparison before and after applying MIAM, including base class accuracy, new class accuracy, final task accuracy, and average accuracy on CIFAR100 and ImageNet-R. Specifically, $m=0$ denotes training without the margin penalty, $m=0.2$ indicates training with a margin penalty of 0.2, and $merge$ refers to using MIAM to merge the two adapter sets learned independently.

Figure \ref{f_balance_c} shows the results on CIFAR100. When using a plain classification loss without a margin penalty ($m=0$) to focus on new class generalization, the model achieves a base class accuracy of 92.5\% and a new class accuracy of 81.1\%. In contrast, applying a large margin penalty ($m=0.2$), which emphasizes base class discriminability, improves base class accuracy to 93.5\%, but significantly reduces new class accuracy to 73.8\%. This trade-off highlights the opposing trends between base class discriminability and new class generalization. By applying the MIAM mechanism, the base class accuracy reaches 93.4\% (a improvement of 0.9\% over $m=0$), while new class accuracy increases to 80.2\% (a substantial improvement of 6.4\% over $m=0.2$). This balanced performance effectively prevents the model from becoming biased toward either base or new classes. Consequently, as shown in the second row of Figure \ref{f_balance_c}, the merged model achieves the highest final and average accuracy, demonstrating the effectiveness of the proposed MIAM mechanism.

Figure \ref{f_balance_r} presents similar results on ImageNet-R. Without a margin penalty ($m=0$), the model obtains 85.3\% base class accuracy and 63.8\% new class accuracy. When using a large margin penalty ($m=0.2$), base class accuracy increases to 87.4\%, but new class accuracy declines to 59.5\%. With MIAM, base class accuracy reaches 86.9\% (a 1.6\% gain over $m=0$), and new class accuracy improves to 64.4\%, which is the highest among all settings and represents a 4.9\% improvement over $m=0.2$. These results are consistent with those on CIFAR100, confirming that MIAM effectively maintains base class performance while significantly enhancing generalization to new classes. Overall, our MIAM mechanism exhibits stronger forward compatibility and achieves a better balance between stability and plasticity.

\begin{figure}[htbp]
\centering
\includegraphics[width=0.48\textwidth]{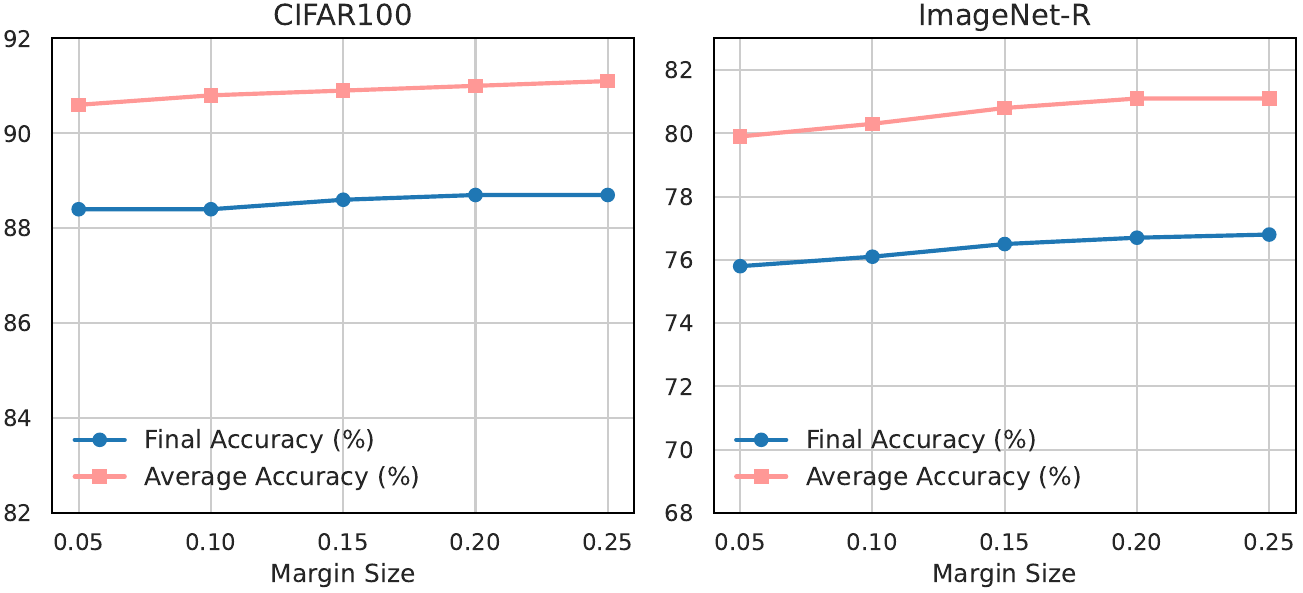}
\caption{\small Effect of margin size on final accuracy $A_{final}$ and average accuracy $A_{avg}$ on CIFAR100 and TinyImageNet.}
\label{f_m_size}
\end{figure}

\noindent \textbf{Impact of Margin Size.}
In Figure \ref{f_m_size}, we evaluate the performance of our SMP method under varying margin sizes on the CIFAR100 and ImageNet-R datasets. The results show that as the margin size $m$ increases from 0.05 to 0.20, both the final and average accuracies steadily improve. However, further increases $m$ to $0.25$, the performance gain becomes less pronounced, suggesting that excessively widening the embeddings and decision boundaries between classes provides limited additional benefits. Overall, $m=0.2$ is a robust choice.

\begin{table}[]
\centering
\caption{Effect of the shot number on final accuracy $A_{final}$ and average accuracy $A_{avg}$ on three benchmark datasets.}
\renewcommand\tabcolsep{1.5mm}
\resizebox{0.9\linewidth}{!}{
\begin{tabular}{ccccccc}
\toprule
\multirow{2}{*}{Method} & \multicolumn{2}{c}{CIFAR100} & \multicolumn{2}{c}{ImageNet-R} & \multicolumn{2}{c}{CUB200} \\ 
                         & $A_{final}$         & $A_{avg}$          & $A_{final}$          & $A_{avg}$           & $A_{final}$        & $A_{avg}$         \\ \midrule
1-shot                   & 79.3          & 85.9         & 65.2           & 74.2          & 77.0         & 81.4        \\
5-shot                   & 88.7          & 91.0         & 76.7           & 81.1          & 86.6         & 87.1        \\
10-shot                  & 89.7          & 91.7         & 78.8           & 82.3          & 88.2         & 88.2        \\
20-shot                  & 90.2          & 92.0         & 79.4           & 82.8          & 88.8         & 88.6        \\ \bottomrule
\end{tabular}}
\label{shot_size}
\end{table}

\noindent \textbf{Impact of Shot Number.}  
In Table \ref{shot_size}, we evaluate the impact of varying shot numbers on the performance of our SMP method across three benchmark datasets. The results show that as the shot number increases, our method consistently achieves better performance. Specifically, on CIFAR-100, the final accuracy improves significantly from 79.3\% (1-shot) to 90.2\% (20-shot), while the average accuracy rises from 85.9\% to 92.0\%. Similar trends are observed on ImageNet-R and CUB-200.

\begin{figure*}[htb]
\centering
\includegraphics[width=\textwidth]{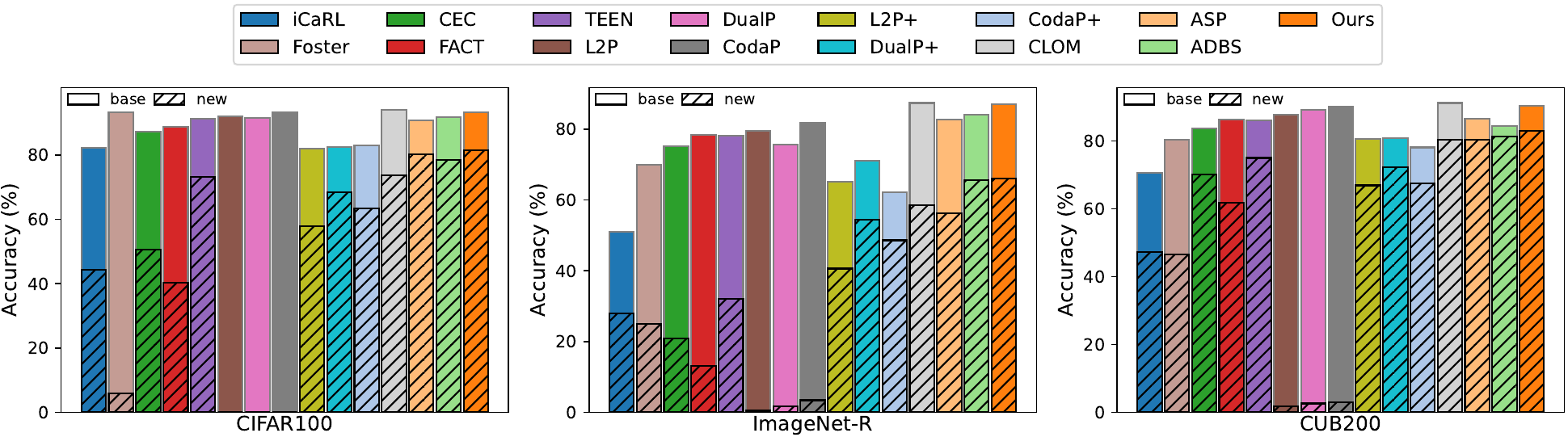}
\caption{\small Detailed comparison of existing methods and SMP in terms of base and new class accuracy at the final incremental task. SMP outperforms all other methods on new classes while maintaining competitive performance on base classes.}
\label{f_base_new}
\end{figure*}

\noindent \textbf{Base and New Class Accuracy.}
In Figure \ref{f_base_new}, we report the detailed accuracy on base classes (from task 0) and new classes (from tasks $t > 0$) on the final task $T$. The results show that our SMP method not only maintains high accuracy on base classes, but also achieves superior performance on new classes. This well-balanced performance across both base and new classes underscores SMP’s capability to retain prior knowledge while exhibiting strong plasticity for adapting to few-shot new classes in subsequent incremental tasks.

\begin{table}[]
\centering
\caption{Ablation study of each component in our method based on a ViT model pre-trained on ImageNet-21K. Final accuracy $A_{final}$ and average accuracy $A_{avg}$ are reported on three benchmark datasets.}
\renewcommand\tabcolsep{2.0mm}
\resizebox{1.0\linewidth}{!}{
\begin{tabular}{cccccccc}
\toprule
\multirow{2}{*}{MIAM} & \multirow{2}{*}{MPCC} & \multicolumn{2}{c}{CIFAR100} & \multicolumn{2}{c}{ImageNet-R} & \multicolumn{2}{c}{CUB200} \\
                 &                            & $A_{final}$         & $A_{avg}$          & $A_{final}$          & $A_{avg}$           & $A_{final}$        & $A_{avg}$         \\ \midrule
                 &                            & 87.7          & 90.3         & 69.5           & 76.4          & 84.9         & 85.3        \\
\checkmark       &                            & 88.4          & 91.0         & 74.1           & 79.6          & 85.6         & 86.2        \\
                 & \checkmark                 & 88.6          & 90.7         & 70.6           & 76.8          & 85.6         & 85.7        \\
\checkmark       & \checkmark                 & 89.2          & 91.4         & 75.1           & 80.0          & 87.0         & 87.1        \\ \bottomrule
\end{tabular}}
\label{abla_i21k}
\end{table}

\noindent \textbf{Effect on Different Pre-Trained Models.}
To further validate the robustness of our SMP method across different pre-trained models, we integrate it into a ViT model pre-trained on ImageNet-21K and perform the same ablation study. As shown in Table \ref{abla_i21k}, each component of SMP shows consistent improvements across CIFAR100, ImageNet-R, and CUB-200, similar to the results observed with the ImageNet-1K pre-trained ViT in Table \ref{abla}. When both MIAM and MPCC components are combined, SMP achieves notable improvements: final accuracy increases by 1.5\%, 5.6\%, and 2.1\%, with corresponding average accuracy gains of 1.1\%, 3.6\%, and 1.8\% on the three datasets, respectively. These results robustly demonstrate SMP's adaptability to different pre-trained models.

\noindent \textbf{Detailed Effect of Classifier Calibration.}
To further evaluate the detailed effect of the MPCC strategy, we analyze prediction results from a decoupling perspective, following the approach proposed in TEEN \cite{fcil_teen}. Specifically, we regard all base classes as positive classes and all new classes as negative classes, thereby converting the FSCIL task into a binary classification problem. Based on the confusion matrix obtained after the final incremental task, we compute the False Negative Rate (FNR) and False Positive Rate (FPR) to quantitatively assess the classifier bias. The definitions of FNR and FPR are given as $\mathrm{FNR}=\frac{\mathrm{FN}}{\mathrm{TP}+\mathrm{FN}} \times 100 \%, \mathrm{FPR}=\frac{\mathrm{FP}}{\mathrm{FP}+\mathrm{TN}} \times 100 \%$, where True Positives (TP) refer to the number of correctly predicted positive instances, False Positives (FP) refer to the number of negative instances incorrectly predicted as positive, True Negatives (TN) refer to the number of correctly predicted negative instances, and False Negatives (FN) refer to the number of positive instances incorrectly predicted as negative.

As shown in Table \ref{cab}, without MPCC (first row), the FPR is substantially higher than the FNR in all three datasets, indicating a high tendency to misclassify new classes as base classes. With MPCC enabled (second row), the FPR consistently decreases across all datasets: from 15.1\% to 13.7\% on CIFAR100, from 16.3\% to 15.2\% on ImageNet-R, and from 2.8\% to 2.0\% on CUB-200. Although a slight increase in FNR is observed, the increase is relatively minor. These results demonstrate that MPCC effectively mitigates the misclassification of new classes into base classes. This improvement can be attributed to the calibration strategy’s ability to adjust inter-class decision boundaries, thereby enhancing overall classification accuracy.

\begin{table}[]
\centering
\caption{False negative rate (FNR, \%) and false positive rate (FPR, \%) on three benchmark datasets.}
\renewcommand\tabcolsep{2.0mm}
\resizebox{0.85\linewidth}{!}{
\begin{tabular}{ccccccc}
\toprule
\multirow{2}{*}{Method} & \multicolumn{2}{c}{CIFAR100} & \multicolumn{2}{c}{ImageNet-R} & \multicolumn{2}{c}{CUB200} \\
                         & FNR          & FPR           & FNR           & FPR            & FNR          & FPR         \\ \midrule
w/o MPCC             & 1.7          & 15.1          & 1.8           & 16.3           & 0.9          & 2.8         \\
w/ MPCC           & 1.9          & 13.7          & 2.0           & 15.2           & 1.1          & 2.0         \\ \bottomrule 
\end{tabular}}
\label{cab}
\end{table}

\noindent \textbf{Effect of Margin Penalty on Classifier Calibration.}
To clarify the contribution of the margin penalty in our MPCC strategy, we conducted ablation studies on the CIFAR100 and ImageNet-R datasets, as shown in Table~\ref{tab_cab_m}. Compared to the uncalibrated method, applying our calibration strategy without the margin penalty yielded a 0.1\% improvement in both final and average accuracy on CIFAR100, and led to a 0.4\% gain in final accuracy and a 0.2\% increase in average accuracy on ImageNet-R. When the margin penalty was incorporated into the calibration process, further improvements were observed: on CIFAR100, final accuracy increased by 0.5\% and average accuracy by 0.2\%; on ImageNet-R, final accuracy improved by 0.5\% and average accuracy by 0.3\%. These results underscore the critical role of the margin penalty in enhancing the effectiveness of our MPCC strategy.

\begin{table}[]
\centering
\caption{Ablation study to evaluate the effect of the margin penalty on the classifier calibration strategy on CIFAR100 and ImageNet-R.}
\renewcommand\tabcolsep{2.0mm}
\resizebox{1.0\linewidth}{!}{
\begin{tabular}{ccccccc}
\toprule
\multirow{2}{*}{Method}   & \multicolumn{2}{c}{CIFAR100} & \multicolumn{2}{c}{ImageNet-R}  \\
                           & $A_{final}$          & $A_{avg}$           & $A_{final}$           & $A_{avg}$       \\ \midrule
No Calibration             & 88.1          & 90.7          & 75.8           & 80.6          \\
Calibrated (w/o Margin)    & 88.2          & 90.8          & 76.2           & 80.8          \\
Calibrated (w/ Margin)     & 88.7          & 91.0          & 76.7           & 81.1          \\ \bottomrule 
\end{tabular}}
\label{tab_cab_m}
\end{table}

\begin{figure*}[htbp]
\centering
\captionsetup[subfigure]{font=footnotesize, labelfont={footnotesize}} 
\hfill
\subfloat[\scriptsize Fisher Magnitudes]{\includegraphics[width=0.48\textwidth]{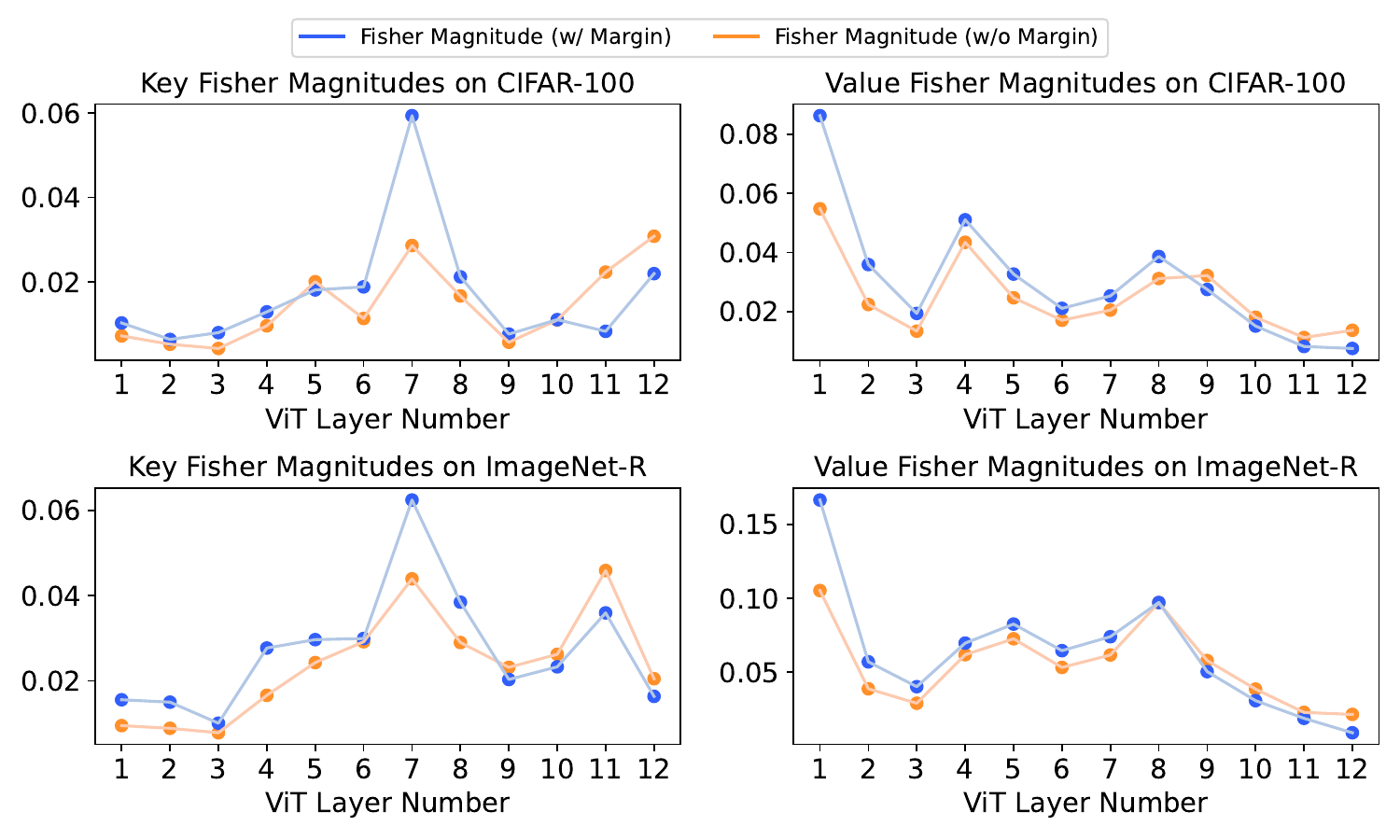}
\label{fig_fish_m}}
\hspace{0px}
\subfloat[\scriptsize Merging Weights]{\includegraphics[width=0.48\textwidth]{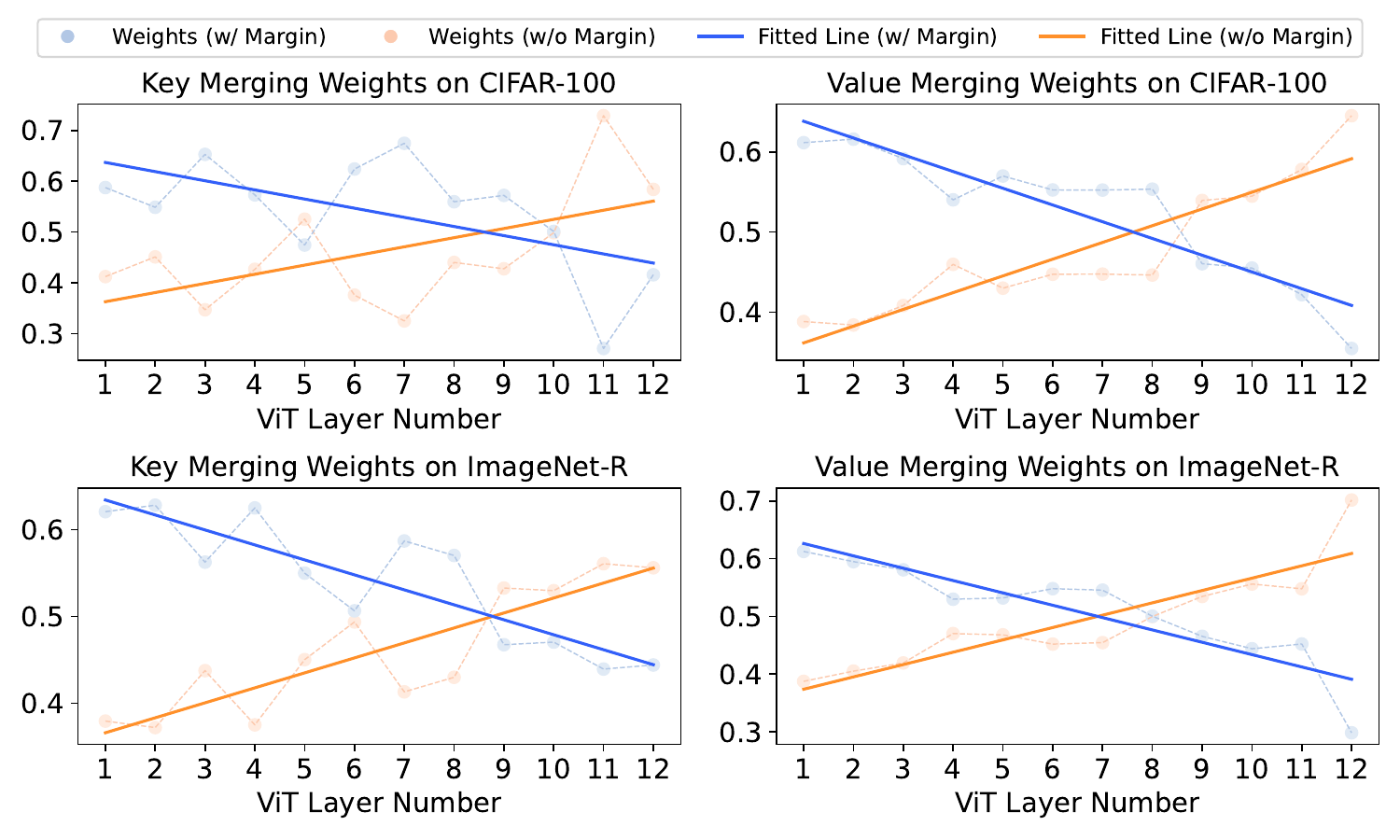}
\label{fig_merge_w}}
\hfill
\caption{\small Fisher magnitudes and merging weights across transformer layers in models with and without margin penalty, evaluated on CIFAR-100 and ImageNet-R. Results are reported for key and value projection matrices in the self-attention module.}
\label{fig_reason}
\end{figure*}

\subsection{Discussion}
We summarize the reasons behind the effectiveness of our adapter merging strategy from the following three perspectives:

\noindent \textbf{(1) Parameter-Efficient Fine-Tuning Foundation.}
Our MIAM mechanism fine-tunes only a small portion of the ViT parameters while keeping the backbone largely frozen. This design not only preserves the general representational capacity of the pre-trained model but also facilitates faster convergence toward FSCIL-specific objectives. Furthermore, updating fewer parameters helps reduce the risk of interference that could occur when directly merging large-scale model parameters.

\noindent \textbf{(2) Adapter Similarity via Intra-Task Optimization.} As shown in Figure \ref{fig_fish_m}, we compare the Fisher magnitude of each basic block for two models trained with and without a margin penalty, which are averaged over the entire training dataset. These comparisons are performed separately on the key and value projection matrices of the self-attention module. Results on CIFAR100 and ImageNet-R showed that, despite minor differences in Fisher magnitudes across layers on key projection, the overall trends are highly consistent under both optimization settings. The Fisher magnitudes on value projection exhibit even stronger consistency. These observations suggest that both sets of adapters, although trained under different loss formulations, maintain strong similarity. This is primarily because they are both optimized on the same base task and share the goal of recognizing the same base classes, leading to aligned optimization directions and feature spaces. Such alignment helps minimize interference during merging and enables the merged adapter to effectively balance performance across both base and new classes.

\noindent \textbf{(3) Adaptive Merging Weights Capture Subtle Differences.}
In Figure \ref{fig_merge_w}, we show the adaptive merging weights for each basic block across the two adapter sets. To better visualize the trend across transformer layers, we apply least squares fitting to obtain smoothed lines. The results indicate that, from shallow to deeper layers, the weights assigned to the margin penalty-based adapters exhibit an approximately gradual decreasing trend, while those assigned to the non-margin adapters show an approximately gradual increasing trend. These trends align closely with the subtle variations in Fisher magnitudes shown in Figure~\ref{fig_fish_m}. Importantly, all merging weights remain within a moderate range (0.3 to 0.7) without drastic fluctuations, further confirming the high similarity between the two adapter sets. Consequently, MIAM can effectively merge their complementary strengths with minimal conflict, yielding a unified adapter set capable of robust performance on both base and new classes.

\section{Conclusion}
We introduce a novel FSCIL method SMP that strategically incorporates margin penalty at different tasks to enhance incremental learning efficiency within the parameter-efficient fine-tuning paradigm. Our method employs two independent sets of low-rank adapters during base task training: one is optimized with a large margin penalty to improve base class discriminability, while the other is trained without margin constraints to facilitate generalization to future new classes. These adapter sets are then adaptively merged using their relative Fisher importance scores, promoting forward compatibility. In addition, we introduce a margin penalty-based classifier calibration strategy to alleviate decision boundary ambiguity in later incremental tasks. Extensive experiments demonstrate that SMP achieves state-of-the-art performance on three FSCIL benchmarks. We believe our method provides valuable insights for advancing forward-compatible learning in class-incremental settings.

\bibliographystyle{IEEEtran}
\bibliography{ref}

\begin{thebibliography}{10}
\providecommand{\url}[1]{#1}
\csname url@samestyle\endcsname
\providecommand{\newblock}{\relax}
\providecommand{\bibinfo}[2]{#2}
\providecommand{\BIBentrySTDinterwordspacing}{\spaceskip=0pt\relax}
\providecommand{\BIBentryALTinterwordstretchfactor}{4}
\providecommand{\BIBentryALTinterwordspacing}{\spaceskip=\fontdimen2\font plus
\BIBentryALTinterwordstretchfactor\fontdimen3\font minus \fontdimen4\font\relax}
\providecommand{\BIBforeignlanguage}[2]{{%
\expandafter\ifx\csname l@#1\endcsname\relax
\typeout{** WARNING: IEEEtran.bst: No hyphenation pattern has been}%
\typeout{** loaded for the language `#1'. Using the pattern for}%
\typeout{** the default language instead.}%
\else
\language=\csname l@#1\endcsname
\fi
#2}}
\providecommand{\BIBdecl}{\relax}
\BIBdecl

\bibitem{cls_surv}
M.~Masana, X.~Liu, B.~Twardowski, M.~Menta, A.~D. Bagdanov, and J.~Van De~Weijer, ``Class-incremental learning: survey and performance evaluation on image classification,'' \emph{IEEE Transactions on Pattern Analysis and Machine Intelligence}, vol.~45, no.~5, pp. 5513--5533, 2022.

\bibitem{cls_surv2}
D.-W. Zhou, Q.-W. Wang, Z.-H. Qi, H.-J. Ye, D.-C. Zhan, and Z.~Liu, ``Class-incremental learning: A survey,'' \emph{IEEE Transactions on Pattern Analysis and Machine Intelligence}, 2024.

\bibitem{cil_vt_lw}
Z.~Tao, L.~Yu, H.~Yao, S.~Huang, and C.~Xu, ``Class incremental learning for light-weighted networks,'' \emph{IEEE Transactions on Circuits and Systems for Video Technology}, 2024.

\bibitem{cil_pass}
F.~Zhu, X.-Y. Zhang, C.~Wang, F.~Yin, and C.-L. Liu, ``Prototype augmentation and self-supervision for incremental learning,'' in \emph{Proceedings of the IEEE/CVF Conference on Computer Vision and Pattern Recognition}, 2021, pp. 5871--5880.

\bibitem{cil_ssre}
K.~Zhu, K.~Zheng, R.~Feng, D.~Zhao, Y.~Cao, and Z.-J. Zha, ``Self-organizing pathway expansion for non-exemplar class-incremental learning,'' in \emph{Proceedings of the IEEE/CVF International Conference on Computer Vision}, 2023, pp. 19\,204--19\,213.

\bibitem{cil_praka}
W.~Shi and M.~Ye, ``Prototype reminiscence and augmented asymmetric knowledge aggregation for non-exemplar class-incremental learning,'' in \emph{Proceedings of the IEEE/CVF International Conference on Computer Vision}, 2023, pp. 1772--1781.

\bibitem{cil_vt_rm}
Y.~Hu, Z.~Liang, X.~Liu, Q.~Hou, and M.-M. Cheng, ``Reformulating classification as image-class matching for class incremental learning,'' \emph{IEEE Transactions on Circuits and Systems for Video Technology}, 2024.

\bibitem{cil_vt_mix}
K.~Song, G.~Liang, Z.~Chen, and Y.~Zhang, ``Non-exemplar class-incremental learning by random auxiliary classes augmentation and mixed features,'' \emph{IEEE Transactions on Circuits and Systems for Video Technology}, vol.~34, no.~9, pp. 7830--7843, 2024.

\bibitem{cil_tass}
X.~Liu, J.-T. Zhai, A.~D. Bagdanov, K.~Li, and M.-M. Cheng, ``Task-adaptive saliency guidance for exemplar-free class incremental learning,'' in \emph{Proceedings of the IEEE/CVF Conference on Computer Vision and Pattern Recognition}, 2024, pp. 23\,954--23\,963.

\bibitem{acmap}
T.~Fukuda, H.~Kera, and K.~Kawamoto, ``Adapter merging with centroid prototype mapping for scalable class-incremental learning,'' in \emph{Proceedings of the Computer Vision and Pattern Recognition Conference}, 2025, pp. 4884--4893.

\bibitem{fcil_surv}
S.~Tian, L.~Li, W.~Li, H.~Ran, X.~Ning, and P.~Tiwari, ``A survey on few-shot class-incremental learning,'' \emph{Neural Networks}, vol. 169, pp. 307--324, 2024.

\bibitem{fcil_fact}
D.-W. Zhou, F.-Y. Wang, H.-J. Ye, L.~Ma, S.~Pu, and D.-C. Zhan, ``Forward compatible few-shot class-incremental learning,'' in \emph{Proceedings of the IEEE/CVF Conference on Computer Vision and Pattern Recognition}, 2022, pp. 9046--9056.

\bibitem{fcil_vt_cap}
F.-Y. Liang, Y.-W. Zhan, J.~Liu, C.-Y. Zhang, Z.-D. Chen, X.~Luo, and X.-S. Xu, ``Class-aware prompting for federated few-shot class-incremental learning,'' \emph{IEEE Transactions on Circuits and Systems for Video Technology}, 2025.

\bibitem{fcil_savc}
Z.~Song, Y.~Zhao, Y.~Shi, P.~Peng, L.~Yuan, and Y.~Tian, ``Learning with fantasy: Semantic-aware virtual contrastive constraint for few-shot class-incremental learning,'' in \emph{Proceedings of the IEEE/CVF Conference on Computer Vision and Pattern Recognition}, 2023, pp. 24\,183--24\,192.

\bibitem{fcil_vt_icec}
Y.~Wang, G.~Zhao, and X.~Qian, ``Improved continually evolved classifiers for few-shot class-incremental learning,'' \emph{IEEE Transactions on Circuits and Systems for Video Technology}, vol.~34, no.~2, pp. 1123--1134, 2023.

\bibitem{fcil_teen}
Q.-W. Wang, D.-W. Zhou, Y.-K. Zhang, D.-C. Zhan, and H.-J. Ye, ``Few-shot class-incremental learning via training-free prototype calibration,'' \emph{Advances in Neural Information Processing Systems}, vol.~36, pp. 15\,060--15\,076, 2023.

\bibitem{fcil_vt_prompt}
S.~Li, F.~Liu, L.~Jiac, L.~Li, P.~Chen, X.~Liu, and W.~Ma, ``Prompt-based concept learning for few-shot class-incremental learning,'' \emph{IEEE Transactions on Circuits and Systems for Video Technology}, 2025.

\bibitem{fcil_clom}
Y.~Zou, S.~Zhang, Y.~Li, and R.~Li, ``Margin-based few-shot class-incremental learning with class-level overfitting mitigation,'' \emph{Advances in Neural Information Processing Systems}, vol.~35, pp. 27\,267--27\,279, 2022.

\bibitem{asp}
C.~Liu, Z.~Wang, T.~Xiong, R.~Chen, Y.~Wu, J.~Guo, and H.~Huang, ``Few-shot class incremental learning with attention-aware self-adaptive prompt,'' in \emph{European Conference on Computer Vision}.\hskip 1em plus 0.5em minus 0.4em\relax Springer, 2024, pp. 1--18.

\bibitem{vit}
A.~Dosovitskiy, L.~Beyer, A.~Kolesnikov, D.~Weissenborn, X.~Zhai, T.~Unterthiner, M.~Dehghani, M.~Minderer, G.~Heigold, S.~Gelly \emph{et~al.}, ``An image is worth 16x16 words: Transformers for image recognition at scale,'' \emph{arXiv preprint arXiv:2010.11929}, 2020.

\bibitem{cifar}
A.~Krizhevsky, G.~Hinton \emph{et~al.}, ``Learning multiple layers of features from tiny images,'' 2009.

\bibitem{imr}
D.~Hendrycks, S.~Basart, N.~Mu, S.~Kadavath, F.~Wang, E.~Dorundo, R.~Desai, T.~Zhu, S.~Parajuli, M.~Guo \emph{et~al.}, ``The many faces of robustness: A critical analysis of out-of-distribution generalization,'' in \emph{Proceedings of the IEEE/CVF International Conference on Computer Vision}, 2021, pp. 8340--8349.

\bibitem{local_merge}
K.~Wang, N.~Dimitriadis, G.~Ortiz-Jimenez, F.~Fleuret, and P.~Frossard, ``Localizing task information for improved model merging and compression,'' \emph{arXiv preprint arXiv:2405.07813}, 2024.

\bibitem{emr_merge}
C.~Huang, P.~Ye, T.~Chen, T.~He, X.~Yue, and W.~Ouyang, ``Emr-merging: Tuning-free high-performance model merging,'' \emph{Advances in Neural Information Processing Systems}, vol.~37, pp. 122\,741--122\,769, 2024.

\bibitem{arith}
G.~Ilharco, M.~T. Ribeiro, M.~Wortsman, S.~Gururangan, L.~Schmidt, H.~Hajishirzi, and A.~Farhadi, ``Editing models with task arithmetic,'' \emph{arXiv preprint arXiv:2212.04089}, 2022.

\bibitem{meta_p1}
G.~Zheng and A.~Zhang, ``Few-shot class-incremental learning with meta-learned class structures,'' in \emph{2021 International Conference on Data Mining Workshops (ICDMW)}.\hskip 1em plus 0.5em minus 0.4em\relax IEEE, 2021, pp. 421--430.

\bibitem{meta_p2}
K.~Chen and C.-G. Lee, ``Incremental few-shot learning via vector quantization in deep embedded space,'' in \emph{International Lonference on Learning Representations}, 2021.

\bibitem{meta_p4}
Y.~Yang, H.~Yuan, X.~Li, Z.~Lin, P.~Torr, and D.~Tao, ``Neural collapse inspired feature-classifier alignment for few-shot class incremental learning,'' \emph{International Conference on Learning Representations}, 2023.

\bibitem{meta_m1}
Z.~Chi, L.~Gu, H.~Liu, Y.~Wang, Y.~Yu, and J.~Tang, ``Metafscil: A meta-learning approach for few-shot class incremental learning,'' in \emph{Proceedings of the IEEE/CVF Conference on Computer Vision and Pattern Recognition}, 2022, pp. 14\,166--14\,175.

\bibitem{dynamic_1}
X.~Tao, X.~Hong, X.~Chang, S.~Dong, X.~Wei, and Y.~Gong, ``Few-shot class-incremental learning,'' in \emph{Proceedings of the IEEE/CVF Conference on Computer Vision and Pattern Recognition}, 2020, pp. 12\,183--12\,192.

\bibitem{dynamic_cec}
C.~Zhang, N.~Song, G.~Lin, Y.~Zheng, P.~Pan, and Y.~Xu, ``Few-shot incremental learning with continually evolved classifiers,'' in \emph{Proceedings of the IEEE/CVF Conference on Computer Vision and Pattern Recognition}, 2021, pp. 12\,455--12\,464.

\bibitem{dynamic_3}
B.~Yang, M.~Lin, Y.~Zhang, B.~Liu, X.~Liang, R.~Ji, and Q.~Ye, ``Dynamic support network for few-shot class incremental learning,'' \emph{IEEE Transactions on Pattern Analysis and Machine Intelligence}, vol.~45, no.~3, pp. 2945--2951, 2022.

\bibitem{replay_d1}
A.~Kukleva, H.~Kuehne, and B.~Schiele, ``Generalized and incremental few-shot learning by explicit learning and calibration without forgetting,'' in \emph{Proceedings of the IEEE/CVF International Conference on Computer Vision}, 2021, pp. 9020--9029.

\bibitem{replay_d2}
A.~Cheraghian, S.~Rahman, P.~Fang, S.~K. Roy, L.~Petersson, and M.~Harandi, ``Semantic-aware knowledge distillation for few-shot class-incremental learning,'' in \emph{Proceedings of the IEEE/CVF Conference on Computer Vision and Pattern Recognition}, 2021, pp. 2534--2543.

\bibitem{replay_g1}
H.~Liu, L.~Gu, Z.~Chi, Y.~Wang, Y.~Yu, J.~Chen, and J.~Tang, ``Few-shot class-incremental learning via entropy-regularized data-free replay,'' in \emph{European Conference on Computer Vision}.\hskip 1em plus 0.5em minus 0.4em\relax Springer, 2022, pp. 146--162.

\bibitem{replay_g2}
A.~Agarwal, B.~Banerjee, F.~Cuzzolin, and S.~Chaudhuri, ``Semantics-driven generative replay for few-shot class incremental learning,'' in \emph{Proceedings of the 30th ACM International Conference on Multimedia}, 2022, pp. 5246--5254.

\bibitem{feat_1}
D.~Wijaya, A.~F. Aky{\"u}rek, E.~Akyurek, and J.~Andreas, ``Subspace regularizers for few-shot class incremental learning,'' 2022.

\bibitem{feat_2}
D.-Y. Kim, D.-J. Han, J.~Seo, and J.~Moon, ``Warping the space: Weight space rotation for class-incremental few-shot learning,'' in \emph{The Eleventh International Conference on Learning Representations}, 2023.

\bibitem{feat_limit}
D.-W. Zhou, H.-J. Ye, L.~Ma, D.~Xie, S.~Pu, and D.-C. Zhan, ``Few-shot class-incremental learning by sampling multi-phase tasks,'' \emph{IEEE Transactions on Pattern Analysis and Machine Intelligence}, vol.~45, no.~11, pp. 12\,816--12\,831, 2022.

\bibitem{peft_survey}
D.-W. Zhou, Z.-W. Cai, H.-J. Ye, D.-C. Zhan, and Z.~Liu, ``Revisiting class-incremental learning with pre-trained models: Generalizability and adaptivity are all you need,'' \emph{International Journal of Computer Vision}, vol. 133, no.~3, pp. 1012--1032, 2025.

\bibitem{l2p}
Z.~Wang, Z.~Zhang, C.-Y. Lee, H.~Zhang, R.~Sun, X.~Ren, G.~Su, V.~Perot, J.~Dy, and T.~Pfister, ``Learning to prompt for continual learning,'' in \emph{Proceedings of the IEEE/CVF Conference on Computer Vision and Pattern Recognition}, 2022, pp. 139--149.

\bibitem{dual}
Z.~Wang, Z.~Zhang, S.~Ebrahimi, R.~Sun, H.~Zhang, C.-Y. Lee, X.~Ren, G.~Su, V.~Perot, J.~Dy \emph{et~al.}, ``Dualprompt: Complementary prompting for rehearsal-free continual learning,'' in \emph{European Conference on Computer Vision}.\hskip 1em plus 0.5em minus 0.4em\relax Springer, 2022, pp. 631--648.

\bibitem{coda}
J.~S. Smith, L.~Karlinsky, V.~Gutta, P.~Cascante-Bonilla, D.~Kim, A.~Arbelle, R.~Panda, R.~Feris, and Z.~Kira, ``Coda-prompt: Continual decomposed attention-based prompting for rehearsal-free continual learning,'' in \emph{Proceedings of the IEEE/CVF Conference on Computer Vision and Pattern Recognition}, 2023, pp. 11\,909--11\,919.

\bibitem{inflora}
Y.-S. Liang and W.-J. Li, ``Inflora: Interference-free low-rank adaptation for continual learning,'' in \emph{Proceedings of the IEEE/CVF Conference on Computer Vision and Pattern Recognition}, 2024, pp. 23\,638--23\,647.

\bibitem{lora_sub}
X.~Liu and X.~Chang, ``Lora subtraction for drift-resistant space in exemplar-free continual learning,'' in \emph{Proceedings of the Computer Vision and Pattern Recognition Conference}, 2025, pp. 15\,308--15\,318.

\bibitem{sphere}
W.~Liu, Y.~Wen, Z.~Yu, M.~Li, B.~Raj, and L.~Song, ``Sphereface: Deep hypersphere embedding for face recognition,'' in \emph{Proceedings of the IEEE Conference on Computer Vision and Pattern Recognition}, 2017, pp. 212--220.

\bibitem{cosfact}
H.~Wang, Y.~Wang, Z.~Zhou, X.~Ji, D.~Gong, J.~Zhou, Z.~Li, and W.~Liu, ``Cosface: Large margin cosine loss for deep face recognition,'' in \emph{Proceedings of the IEEE Conference on Computer Vision and Pattern Recognition}, 2018, pp. 5265--5274.

\bibitem{add_margin}
F.~Wang, J.~Cheng, W.~Liu, and H.~Liu, ``Additive margin softmax for face verification,'' \emph{IEEE Signal Processing Letters}, vol.~25, no.~7, pp. 926--930, 2018.

\bibitem{Arcface}
J.~Deng, J.~Guo, N.~Xue, and S.~Zafeiriou, ``Arcface: Additive angular margin loss for deep face recognition,'' in \emph{Proceedings of the IEEE/CVF Conference on Computer Vision and Pattern Recognition}, 2019, pp. 4690--4699.

\bibitem{neg_margin}
B.~Liu, Y.~Cao, Y.~Lin, Q.~Li, Z.~Zhang, M.~Long, and H.~Hu, ``Negative margin matters: Understanding margin in few-shot classification,'' in \emph{Computer Vision--ECCV 2020: 16th European Conference, Glasgow, UK, August 23--28, 2020, Proceedings, Part IV 16}.\hskip 1em plus 0.5em minus 0.4em\relax Springer, 2020, pp. 438--455.

\bibitem{ewc}
J.~Kirkpatrick, R.~Pascanu, N.~Rabinowitz, J.~Veness, G.~Desjardins, A.~A. Rusu, K.~Milan, J.~Quan, T.~Ramalho, A.~Grabska-Barwinska \emph{et~al.}, ``Overcoming catastrophic forgetting in neural networks,'' \emph{Proceedings of the National Academy of Sciences}, vol. 114, no.~13, pp. 3521--3526, 2017.

\bibitem{hide}
L.~Wang, J.~Xie, X.~Zhang, M.~Huang, H.~Su, and J.~Zhu, ``Hierarchical decomposition of prompt-based continual learning: Rethinking obscured sub-optimality,'' \emph{Advances in Neural Information Processing Systems}, vol.~36, pp. 69\,054--69\,076, 2023.

\bibitem{ssiat}
Y.~Tan, Q.~Zhou, X.~Xiang, K.~Wang, Y.~Wu, and Y.~Li, ``Semantically-shifted incremental adapter-tuning is a continual vitransformer,'' in \emph{Proceedings of the IEEE/CVF Conference on Computer Vision and Pattern Recognition}, 2024, pp. 23\,252--23\,262.

\bibitem{cub200}
C.~Wah, S.~Branson, P.~Welinder, P.~Perona, and S.~Belongie, ``The caltech-ucsd birds-200-2011 dataset,'' 2011.

\bibitem{icarl}
S.-A. Rebuffi, A.~Kolesnikov, G.~Sperl, and C.~H. Lampert, ``icarl: Incremental classifier and representation learning,'' in \emph{Proceedings of the IEEE Conference on Computer Vision and Pattern Recognition}, 2017, pp. 2001--2010.

\bibitem{foster}
F.-Y. Wang, D.-W. Zhou, H.-J. Ye, and D.-C. Zhan, ``Foster: Feature boosting and compression for class-incremental learning,'' in \emph{European Conference on Computer Vision}.\hskip 1em plus 0.5em minus 0.4em\relax Springer, 2022, pp. 398--414.

\bibitem{adbs}
L.~Li, Y.~Tan, S.~Yang, H.~Cheng, Y.~Dong, and L.~Yang, ``Adaptive decision boundary for few-shot class-incremental learning,'' in \emph{Proceedings of the AAAI Conference on Artificial Intelligence}, vol.~39, no.~17, 2025, pp. 18\,359--18\,367.

\bibitem{first}
X.~Tao, X.~Hong, X.~Chang, S.~Dong, X.~Wei, and Y.~Gong, ``Few-shot class-incremental learning,'' in \emph{Proceedings of the IEEE/CVF Conference on Computer Vision and Pattern Recognition}, 2020, pp. 12\,183--12\,192.

\bibitem{hacc}
C.~Peng, K.~Zhao, T.~Wang, M.~Li, and B.~C. Lovell, ``Few-shot class-incremental learning from an open-set perspective,'' in \emph{European Conference on Computer Vision}.\hskip 1em plus 0.5em minus 0.4em\relax Springer, 2022, pp. 382--397.

\end{thebibliography}

\end{document}